\documentclass[10pt,twocolumn,letterpaper]{article}

\usepackage{graphicx}
\usepackage{float}
\usepackage{animate}
\usepackage{subcaption}
\usepackage[pagenumbers]{cvpr}

\definecolor{cvprblue}{rgb}{0.21,0.49,0.74}
\usepackage[pagebackref,breaklinks,colorlinks,allcolors=cvprblue]{hyperref}

\def\confName{CVPR}
\def\confYear{2025}

\title{Sketch-Guided Stylized Landscape Cinemagraph Synthesis}

\author{
Hao Jin\textsuperscript{1}, Hengyuan Chang\textsuperscript{1}, Xiaoxuan Xie\textsuperscript{1}, Zhengyang Wang\textsuperscript{1}, Xusheng Du\textsuperscript{1}, \\
Shaojun Hu\textsuperscript{2}, Haoran Xie\textsuperscript{1, 3}\\
\textsuperscript{1}Japan Advanced Institute of Science and Technology (JAIST)\\
\textsuperscript{2}Northwest A\&F University \\
\textsuperscript{3}Waseda University
}

\begin{document}

\maketitle

\begin{abstract}
Designing stylized cinemagraphs is challenging due to the difficulty in customizing complex and expressive flow elements. To achieve intuitive and detailed control of the generated cinemagraphs, sketches provide a feasible solution to convey personalized design requirements beyond text inputs. In this paper, we propose Sketch2Cinemagraph, a sketch-guided framework that enables the conditional generation of stylized cinemagraphs from freehand sketches. Sketch2Cinemagraph adopts text prompts for initial landscape generation and provides sketch controls for both spatial and motion cues. The latent diffusion model first generates target stylized landscape images along with realistic versions. Then, a pre-trained object detection model obtains masks for the flow regions. We propose a latent motion diffusion model to estimate motion field in fluid regions of the generated landscape images. The input motion sketches serve as the conditions to control the generated motion fields in the masked fluid regions with the prompt. To synthesize cinemagraph frames, the pixels within fluid regions are warped to target locations at each timestep using a U-Net based frame generator. The results verified that Sketch2Cinemagraph can generate aesthetically appealing stylized cinemagraphs with continuous temporal flow from sketch inputs. We showcase the advantages of Sketch2Cinemagraph through qualitative and quantitative comparisons against the state-of-the-art approaches.
\end{abstract}

\section{Introduction}
\label{sec:intro}

\begin{figure*}[t]
    \begin{center}
	\centering
	\includegraphics[width=\linewidth]{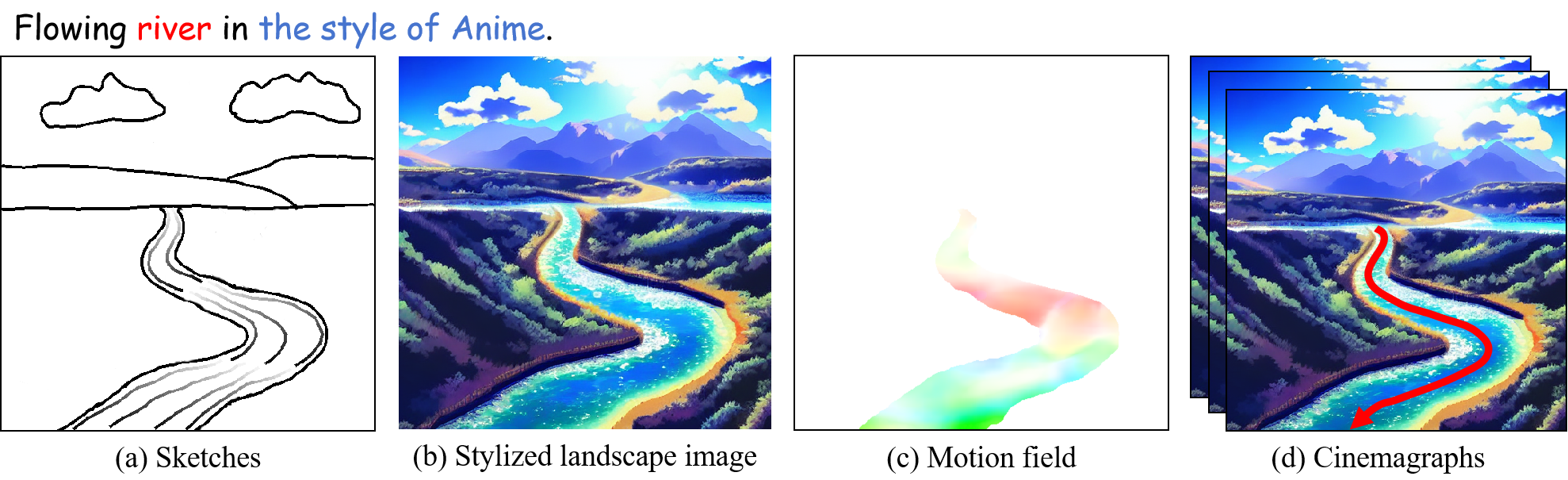}
	\caption{Given the input (a) sketches with motion sketches and text prompt, Sketch2Cinemagraph can synthesize  (d) cinemagraphs from the (b) stylized landscape image with the generated (c) motion fields. The red font represents the text prompt for flow generation, and the blue font for style generation. The white-to-black gradient lines depict the motion sketches for the flow motion direction.}
	\label{fig:teaser}
\end{center}
\end{figure*}

Cinemagraphs embody a captivating fusion of still photography and dynamic video, achieving a unique visual effect by seamlessly integrating motion into static images \cite{Cinemagraphs2011}. As a widely adopted media format, cinemagraphs possess a more vivid and dynamic quality compared to static images. Additionally, cinemagraphs emphasize motion against a static background, providing more captivating and noticeable visual dynamics than normal videos. Nevertheless, generating high-quality cinemagraphs remains a time-consuming and expertise-demanding task. It is challenging for amateurs to create cinemagraphs due to the required skills and experience in image editing and animation design. The traditional approaches for cinemagraph synthesis usually convert from video clips by detecting and extracting dynamic regions from frames while keeping other parts static \cite{yeh2012tool,bai2013automatic, yeh2015selecting, yan2017turning}. Other solutions include the estimation of physical properties of flow elements from static images to infer motion fields \cite{okabe2011creating, jhou2015animating, sugimoto2022water}. All these previous works require reference videos with embedded motions or physically simulated actions. They are limited in motion diversity and struggle to generalize to unseen or diverse scenarios.

Generative models, including generative adversarial networks (GANs) \cite{liao2022text} and diffusion models \cite{rombach2022high}, have attracted significant attention due to their effectiveness in creating captivating high-quality images and videos\cite{choi2024, li2024, bertiche2023}. GAN inversion and deep feature warping were adopted for cinemagraph generation~\cite{choi2024}. 
Text2Cinemagraph~\cite{mahapatra2023} synthesizes artistic-styled cinemagraphs by Eulerian displacement fields warping, leveraging paired realistic and artistic images generated via the latent diffusion model to enhance visual expressiveness. Mahapatra and Kulkarni\cite{mahapatra2022controllable} animated the rasterized fluid elements by approximating the motion from user-provided flow hints. Li et al.\cite{li2024_GenerativeImageDynamics} modeled the long-term motion prior in the Fourier domain, enabling interaction with natural objects. 
These methods can enhance the controllability of cinemagraphs, providing flexibility and precision in manipulating the visual contents. However, they have yet to explore more freeform and intuitive approaches to create cinemagraphs. Current methods are limited to basic motion controls, such as text and arrows, constraining users from fully expressing their creative intents. For content design, sketching is easily accessible and versatile for rapidly prototyping ideas and concepts \cite{xu2022deep}. Various sketch-based methods for image and video generation and editing have been developed \cite{huang2023Anifacedraw, zheng2023sketch, liu2022deepfacevideoediting}. To this end, this paper presents a novel framework that generates fluid elements and their motions from user sketches to create personalized cinemagraphs. 

\begin{figure*}[t]
\centering
\includegraphics[width=\linewidth]{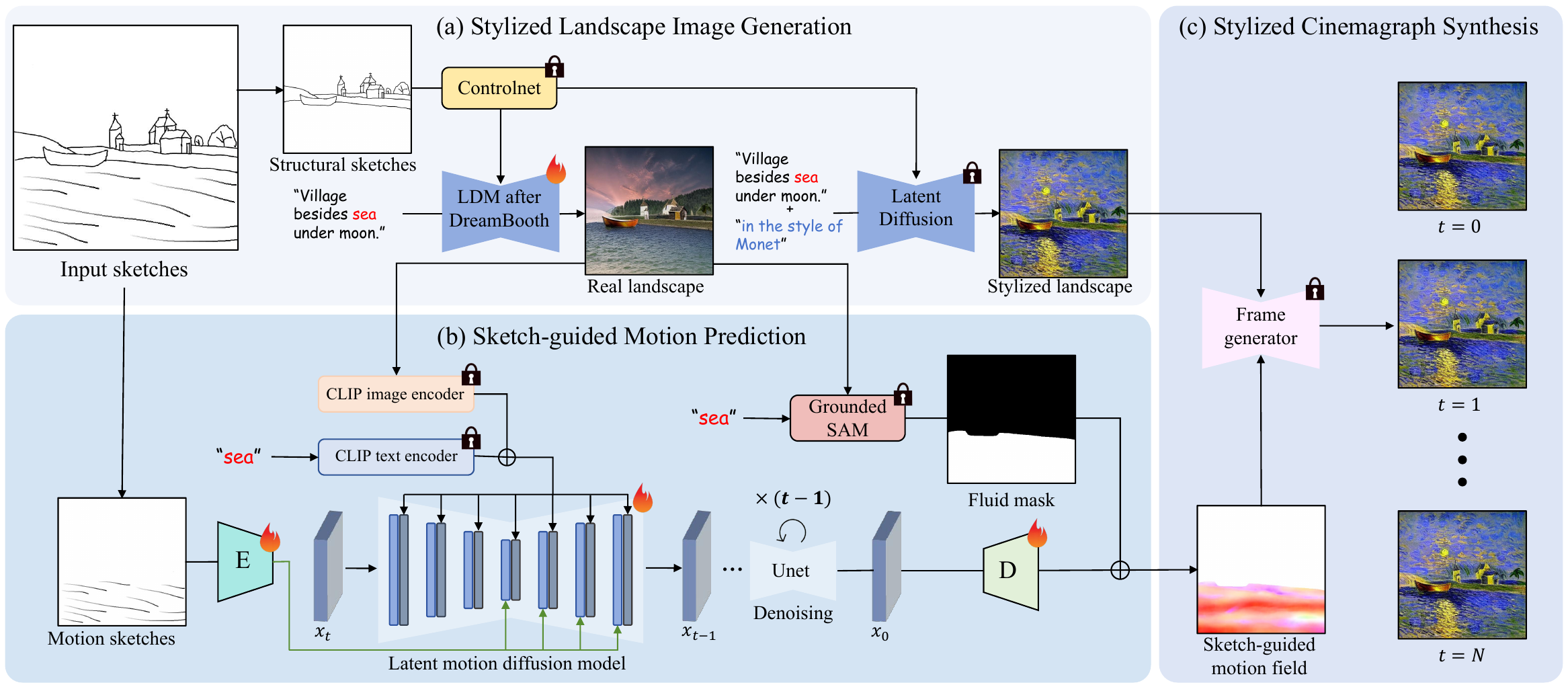}
\caption{The Workflow of \textit{Sketch2Cinemagraph}. Given input hand-drawn sketches of landscape structure and motions, the proposed framework can generate landscape cinemagraphs with (a) stylized landscape image generation, (b) sketch-guided motion prediction, and (c) stylized cinemagraph synthesis stages.}
\label{fig:workflow}
\end{figure*}

In this work, we propose Sketch2Cinemagraph, a sketch-guided framework for generating stylized landscape cinemagraphs from hand-drawn sketches, as shown in Fig. \ref{fig:teaser}. Firstly, the proposed framework generates stylized landscape images incorporating fluid elements, guided by structural sketches and text prompts that specify both fluid dynamics and artistic style. Additionally, it produces corresponding realistic versions that maintain the same spatial structure. Then, the motion field for fluid is estimated from the generated realistic landscape image, conditioned with motion sketches and fluid regions with text prompts using a diffusion model-based motion prediction network. Finally, pixels in the stylized landscape image are warped to their future positions based on the motion field, resulting in visually pleasing cinemagraphs with specified styles. Moreover, our framework exhibits significant flexibility by supporting real-world landscape photographs as direct inputs. It can skips the initial image generation step and estimates the motion field based on provided motion sketches and real-world landscape, allowing users to animate existing photographs into realistic cinemagraphs.

In summary, our main contributions are listed as follows:
\begin{itemize}
\item  A novel framework for landscape cinemagraph synthesis that uses a sketch-guided approach to generate directly from freehand sketches with diffusion models.

\item We introduce the Latent Motion Diffusion Model (LMDM), a diffusion-based network that predicts motion fields of fluid elements in landscape images based on input motion sketches.

\item By validating our Sketch2Cinemagraph against state-of-the-art methods, we show that our approach generates cinemagraphs with more natural and flexible motions.
\end{itemize}

\section{Related Work}

\label{sec:formatting}

\subsection{Animation from Still Images}
The challenge of animating still images is equivalent to simulating or reproducing the physically accurate visual features of natural objects in raster images. Several studies have explored physics-based natural objects simulation, including trees, rivers, clouds, smoke, and animal flocks, to achieve scenery animations \cite{chuang2005animating, xu2008animating, habel2009physically, jhou2015animating, ma2009motion}. Walker et al. \cite{walker2016uncertain} proposed a conditional variational autoencoder to predict the dense trajectory of pixels for future frame prediction. Logacheva et al.\cite{logacheva2020deeplandscape} proposed a fine-tuned StyleGAN to mix realistic still images and time-lapse videos prior for the given photograph animation. Fan et al. \cite{fan2023simulating} simulated 2.5D transparent liquid by integrating physics-based and learning-based methods. Recent studies have incorporated Denoising Diffusion Probabilistic Models (DDPM) \cite{ho2020denoising} into single-image animation tasks. AnimateDiff \cite{guo2023animatediff} proposed a practical pipeline that utilizes personalized text-guided image generative diffusion models for animation generation without specific fine-tuning. However, it struggles with maintaining consistency with real-world physical laws. Zhai et al.\cite{zhai2024anfluid} integrated physics-aware simulation with dual-flow texture learning to improve the performance of natural fluid animation. Our method aims to generate dynamic fluid elements with a continuous flowing structure.

\subsection{Sketch-guided Generative Models}
The sketch-guided generation tasks relevant to our work primarily focus on image and video generation. Chen and Hays\cite{chen2018sketchygan} proposed SketchyGAN, which is trained by augmented paired edge maps and photos to generate lifelike images from freehand sketches. Artistic images can be generated from sketches conditioned on style images \cite{liu2020sketch}. Recently, sketch-guided image generation using diffusion models has achieved significant advancements. Mou et al. \cite{mou2024t2i} introduced T2I-Adapters, a lightweight model designed to provide additional guidance, such as depth, semantic segmentation, and sketches, to text-to-image diffusion models. Voynov et al. \cite{voynov2023sketch} guided a pre-trained text-to-image diffusion model with a spatial map from the sketch domain during inference time. Koley et al. \cite{koley2024s} investigated the potential of sketches in diffusion models and introduce an abstraction-aware framework that allows amateur sketches to produce precise images without the need for textual prompts. Peng et al.\cite{peng2023difffacesketch} proposed a latent diffusion model trained by a two-stage process to achieve high-quality face synthesis. In the context of video generation, Zhang et al. \cite{zhang2021sketch} introduced a two-stage sketch-to-video generation method that allows users to create videos with two rough hand-drawn sketches. Li et al. \cite{li2021deep} matched a sketch with cartoon video frames and used a blending method to create a middle frame guided by the sketches. Instead of interpolating frames, Zheng et al. \cite{zheng2023sketch} created abstract and dynamic sketches using Scalable Vector Graphics (SVG) within the input video, enabling applications of video editing and doodles.

\subsection{Cinemagraph Generation}
Unlike videos, most points in cinemagraphs remain static, with certain elements animating in a seamless loop. Semi-automatic systems have been developed to loop the highlighted region of input videos \cite{tompkin2011towards, joshi2012cliplets, yeh2012tool, bai2013automatic,yan2017turning}. \cite{yeh2012tool, tompkin2011towards, joshi2012cliplets} require user editing, while \cite{yan2017turning, bai2013automatic} struggle with large underlined object movements. To address these issues, Oh et al. \cite{oh2017personalized} proposed a semantic-aware-per-pixel optimization and human preference prediction to create cinemagraphs without user input. Endo et al. \cite{endo2019animating} created high-fidelity long-term waters and skies via convolutional neural networks (CNNs). Holynski et al. \cite{holynski2021animating} predicted the optical flow maps of fluid regions and used deep warping pixels to generate continuous flow. Mahapatra and Kulkarni \cite{mahapatra2022controllable} converted user-specified sparse arrow directions to dense flow via a flow-refinement network to achieve controllable motion prediction. StyleCineGAN \cite{choi2024} produced high-quality landscape cinemagraphs using warping multi-scale deep features encoded by pre-trained StyleGAN. In addition, artistic landscape cinemagraphs are generated by the text-guided diffusion model and optical flow prediction \cite{mahapatra2023}. Li et al. \cite{li20233d} introduced 3D cinemagraphy, elevating 2D motion into 3D to animate 3D landscapes. LoopGaussian \cite{li2024loopgaussian} reconstructed 3D geometry from multi-view photos and inherent scene self-similarity with an Eulerian motion field. Aligning with those works, our work resembles these landscape cinemagraph generation methods using a motion prediction network to estimate the motion fields of flow elements. Furthermore, our approach uniquely generates dynamic elements and a static background directly from simple sketches, thereby achieving full control over both the content and motion levels.

Diffusion models have been successfully applied to motion field synthesis. For example, Ni et al. \cite{ni2023conditional} introduces a Latent Flow Diffusion Model to generate optical flow sequences conditioned on semantic cues such as class labels. This approach effectively synthesizes realistic motions consistent with object categories. Distinct from semantic-guided methods, our work focuses on spatial controllability. The proposed LMDM to address the challenge of sketch-based guidance, where the objective is to translate sparse user-defined strokes into dense fluid motion fields. This formulation prioritizes fine-grained, region-specific directional control, offering a flexible alternative to class-based animation for stylized landscape creation.

\section{Method}
In this work, we propose \textit{Sketch2Cinemagraph}, a sketch-guided generation framework to synthesize the looping stylized landscape cinemagraphs from the input sketches. Fig. \ref{fig:workflow} illustrates the workflow of the proposed sketch-guided cinemagraph synthesis framework. For hand-drawn sketch inputs, we observe that \textbf{structural sketch} and \textbf{motion sketch} can be provided at different stages, allowing independent design of landscape and cinemagraph motion. The motion sketch may use white-to-black gradient lines to depict the flow motion direction, while the structural sketch may employ solid black lines for spatial control. The proposed structure-motion coupling mechanism integrates both structural and motion sketches. By employing these sketches as the unified input for both structure and motion, this mechanism ensures a consistent interaction experience. More crucially, the structural sketch serves as a shared geometric constraint to synchronize the dual-stream generation process. Because the stylized landscape and the realistic reference (utilized for motion inference) are synthesized by separate diffusion processes, their spatial layouts would naturally diverge without explicit guidance. The structural sketch enforces strict structural alignment between the two domains, ensuring that the fluid dynamics derived from realistic features map faithfully onto the stylized image. This geometric synchronization may effectively prevent spatial misalignment, thereby eliminating artifacts such as motion bleeding.

The whole framework of Sketch2Cinemagraph consists of the following stages: stylized landscape image generation, sketch-guided motion prediction, and stylized cinemagraph synthesis. Firstly, in the stage of stylized landscape image generation as illustrated in Fig. \ref{fig:workflow}(a), both stylized and realistic landscape images are synthesized from structural sketches conditioned on text prompts specifying the landscape elements, such as waterfall, sea, and river (Sec. \ref{subsec:LandscapeImageGeneration}). Subsequently, as illustrated in Fig. \ref{fig:workflow}(b), the masks for fluid elements are accurately extracted using the Segment Anything Model (SAM) \cite{kirillov2023segment} and applied to downstream motion field prediction tasks. A diffusion model-based motion prediction network accepts realistic landscape image, text prompt, mask, and user-provided motion sketches to synthesize a sketch-guided motion field, which represents the flow trends of pixels in the fluid regions (Sec. \ref{subsec:motion_field_predict}). Finally, at the stage of stylized cinemagraph synthesis as shown in Fig. \ref{fig:workflow}(c), the looping cinemagraph frames are obtained via Euler-integration and symmetric-splatting (Sec. \ref{subsec:cinemagraph_generation}). 

\begin{figure}[t]
\centering
\includegraphics[width=\linewidth]{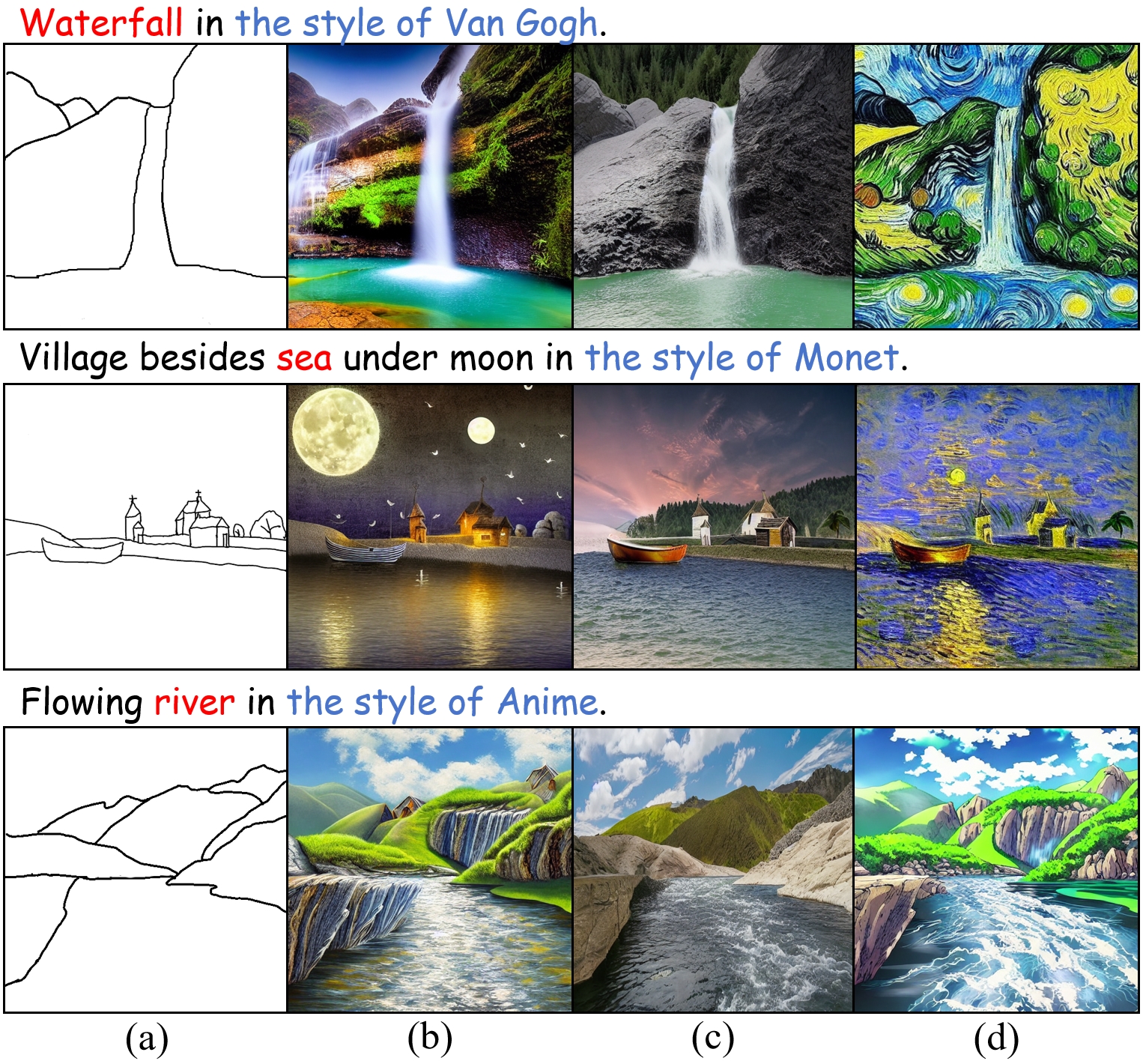}
\caption{Paired landscape image generation using partly fine-tuned ControlNet. The red texts indicate landscape elements, and the blue texts represent the provided style. (a) structural sketches; (b) landscape image generated by pre-trained ControlNet; (c) photo-realistic landscape image generated by partly fine-tuned ControlNet; (d) generated landscape image in specific style by pre-trained ControlNet.}
\label{fig:after_dreambooth}
\end{figure}

\subsection{Preliminary: Text-to-Image Generation}
The latent diffusion model (LDM) \cite{rombach2022high} generates high-quality images by applying the diffusion process in the embedding space rather than in the pixel space. The initial noise map $\epsilon \sim \mathcal N(0, I)$ sampled from a Gaussian distribution and conditioning vector $c$, for example text, sketch or depth map. An image $I$ is generated through gradually removing noise through a denoising U-Net $\epsilon_{\theta}$ in reverse diffusion steps. During training, latent diffusion models aim to minimize a denoising objective function with the loss function $\mathcal{L}$:
\begin{equation}
\label{eqn:ldm_loss}
\mathcal{L}=\mathbb{E}_{x_0, c, \epsilon \sim \mathcal{N}(0,I), t}\left[\left\|\epsilon-\epsilon_{\theta}\left(x_{t}, t, c\right)\right\|_{2}^{2}\right],
\end{equation}
where $x_0$ denotes image latent code, $t$ denotes a timestep uniformly sampled from $1,...,T$, $T$ denotes the number of diffusion steps, and ${x}_0$ denotes the input image.

For spatial conditioning controls, ControlNet~\cite{zhang2023adding} augments large pre-trained text-to-image diffusion models with conditions, including human pose, sketch, and depth. Instead of fine-tuning the entire model, the robust backbone is utilized to learn diverse conditional controls by freezing the deep encoding layers. Adding purpose-specific condition $c_\mathrm{f}$ to the conditioning set, the denoising network $\epsilon_{\theta}$ learns to predict the noise added to the noisy image $x_{t}$. The loss function is designed as follows:
\begin{equation}
\label{eqn:control_loss}
\mathcal{L}=\mathbb{E}_{x_0, c_t,c_\mathrm{f},\epsilon\sim\mathcal{N}(0,1), t}\left[\|\epsilon-\epsilon_\theta(x_t,t,c_t,c_\mathrm{f}))\|_2^2\right],
\end{equation}
where $c_{t}$ denotes text prompts. The model can predict high-quality images by utilizing zero convolutions, which do not add noise to the network. In this paper, a latent diffusion model (LDM) combined with ControlNet is adopted to generate high-quality landscape images conditioned on structural sketches and text prompts.

\subsection{Stylized Landscape Image Generation}
\label{subsec:LandscapeImageGeneration}
The first step of \textit{Sketch2Cinemagraph} aims to generate stylized landscape images and their realistic versions from the given structural sketch and text. In this work, we specifically focused on the scene images containing waterfalls, rivers, and seas. We adopted ControlNet for image generation tasks. This model can generate high-quality landscape images in the provided styles (Fig. \ref{fig:after_dreambooth} (d)). However, we observe that using the pretrained ControlNet poses challenges in generating realistic natural landscape images from sketches, as shown in Fig. \ref{fig:after_dreambooth} (b). Training a ControlNet from scratch for the landscape image generation task would not only require substantial computational resources, but more critically, a large collection of high-quality paired “landscape image–sketch” samples. Constructing such a dataset is prohibitively expensive, as each sketch must accurately reflect the spatial layout and geometric structures of the corresponding landscape image, making manual annotation both time-consuming and labor-intensive. Moreover, even if such a dataset were available, training a new ControlNet from scratch risks discarding the strong sketch-understanding capability already acquired by the pretrained ControlNet.

\begin{figure}[t]
\centering
\includegraphics[width=0.95\linewidth]{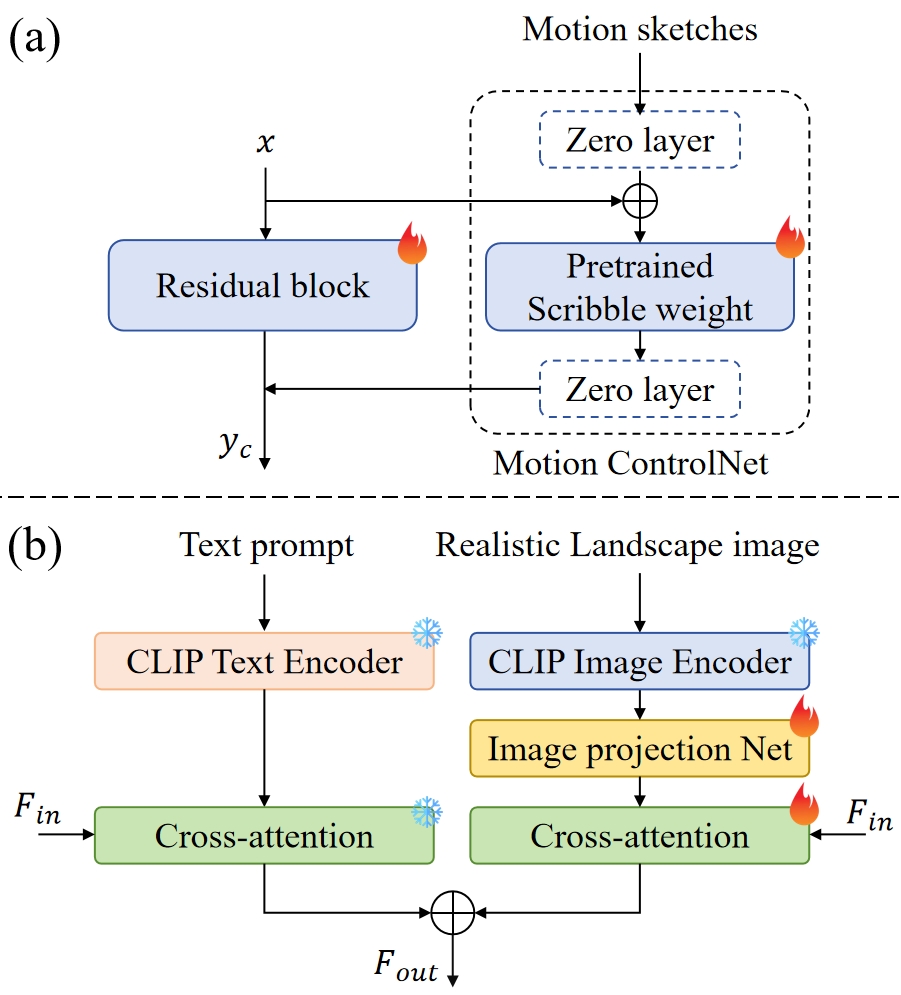}
\caption{(a) ControlNet for motion sketches encoding; (b) Added cross-attention layers for image features. The output $F_{out}$ is fused from text and image embeddings.}
\label{fig:attention}
\end{figure}

\begin{figure}[t]
\centering
\includegraphics[width=\linewidth]{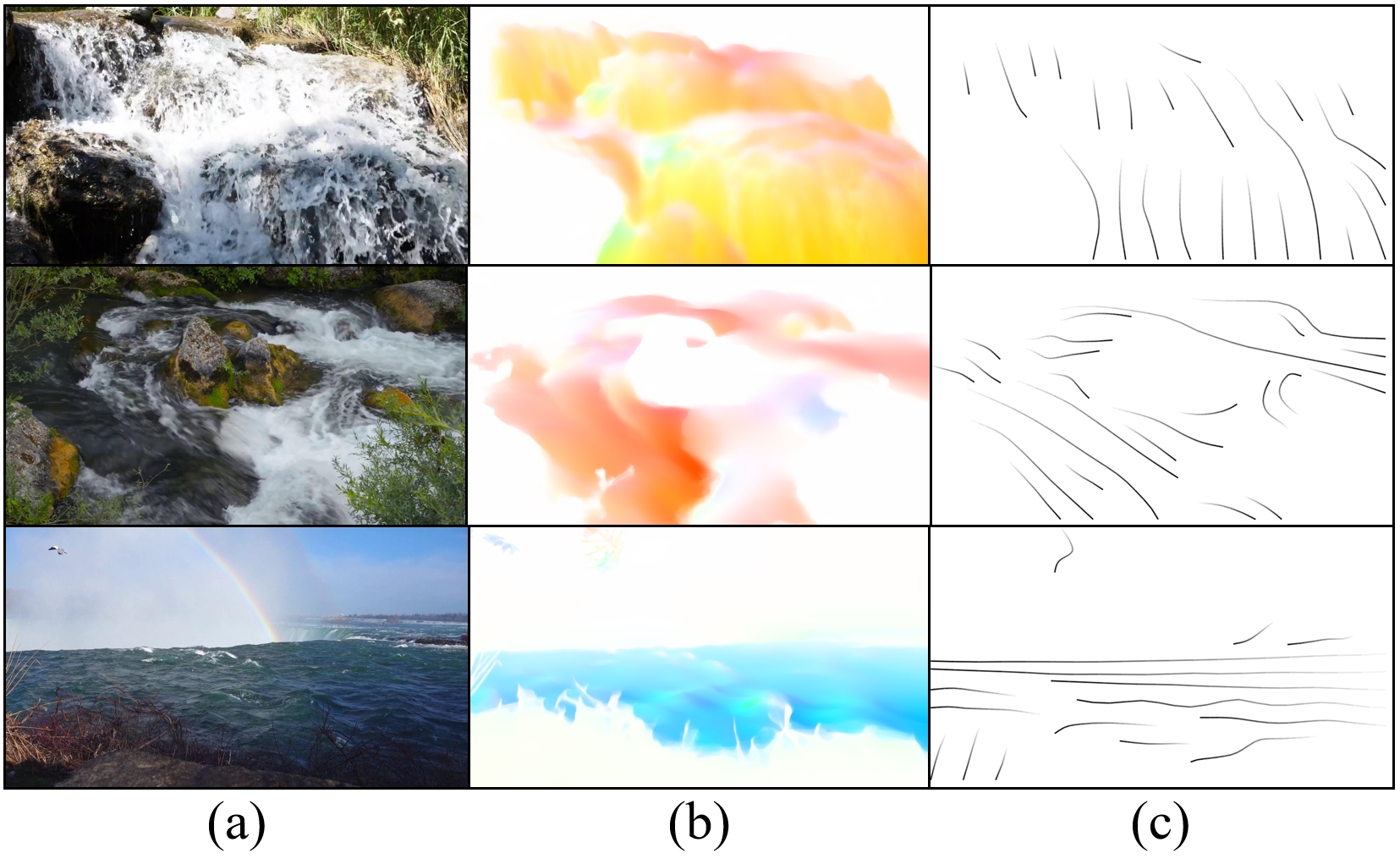}
\caption{Examples in landscape dataset: (a) landscape images; (b) generated motion fields; (c) extracted streamlines from motion fields which can serve as ground truth motion sketches.}
\label{fig:dataset}
\end{figure}

To solve these issues, we decouple the LDM component from the pretrained ControlNet and fine-tune it independently, while keeping the weights of the sketch-conditioned control module unchanged. We adopt DreamBooth \cite{ruiz2023dreambooth}, an image personalization technique, to fine-tune the LDM component on the image subset of the landscape dataset \cite{holynski2021animating}, which contains 4,750 realistic landscape images. We choose DreamBooth for fine-tuning because it offers superior class-level appearance customization, which is essential for accurately representing different types of natural fluids. By fine-tuning on distinct fluid textures (e.g., waterfall, sea, and river), DreamBooth enables the diffusion backbone to capture fine-grained visual cues such as flow patterns, water translucency, and surface turbulence. As a result, the synthesized landscape images exhibit fluid regions that are significantly more visually realistic than those produced by generic models, providing high-quality and stable static inputs for the subsequent motion generation stage. Specifically, we fine-tune the LDM using the natural class labels inherent to our dataset, such as ``river'', ``sea'' and ``waterfall''. This class-level adaptation enables the diffusion model to better capture dataset-specific landscape appearances while maintaining a simple and generalizable text interface. Notably, we exclude the ``smoke'' category from this fine-tuning process, as the pre-trained LDM already demonstrates high fidelity in generating smoke patterns consistent with our dataset. After that, the fine-tuned LDM component is recombined with the original control module to facilitate the generation of photorealistic landscape images conditioned on structural sketches. As shown in Fig. \ref{fig:after_dreambooth} (c), the generated results successfully render photorealistic fluid elements while maintaining structural and compositional consistency.

\begin{figure}[t]
\centering
\includegraphics[width=\linewidth]{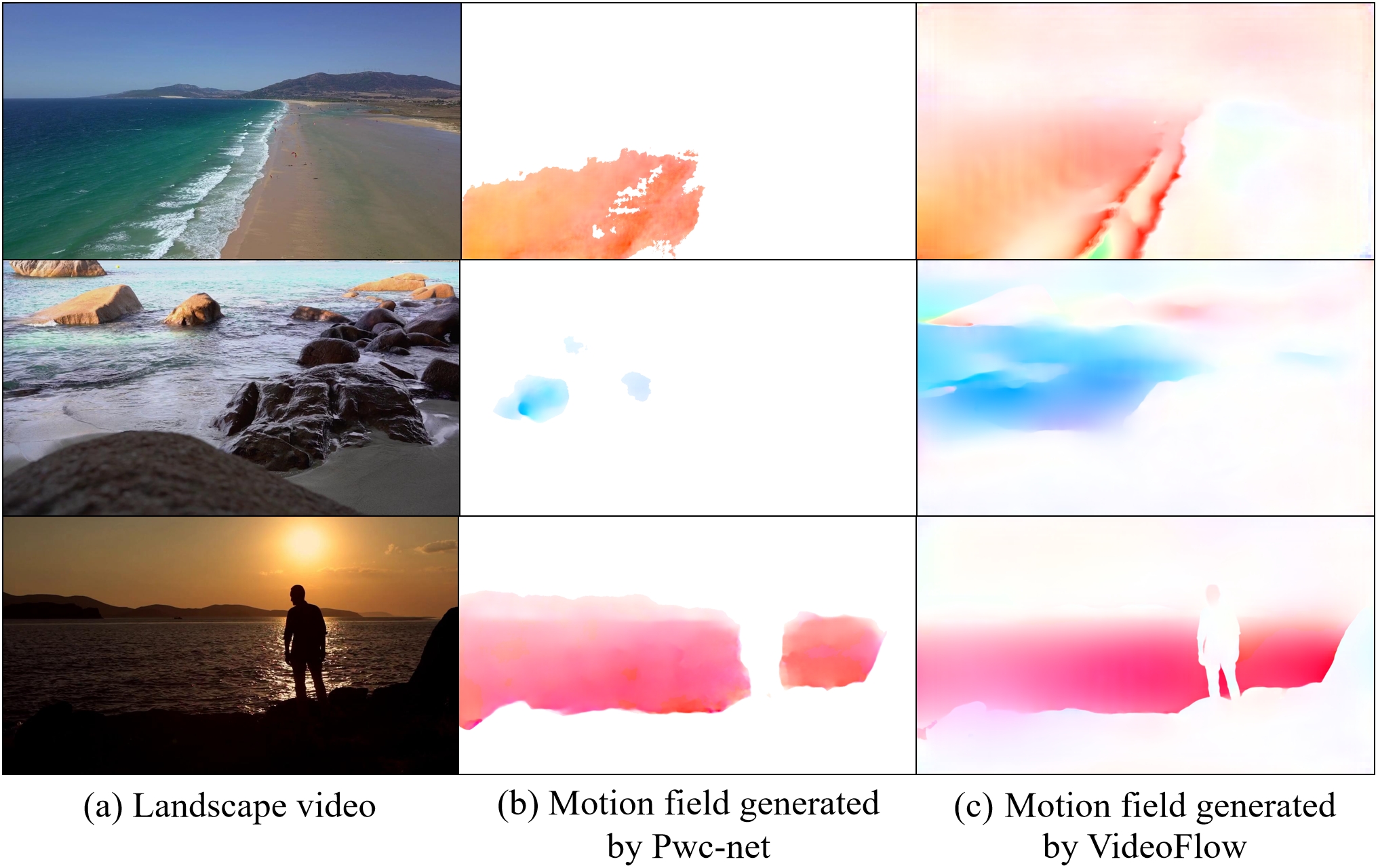}
\caption{Ground truth motion field generation from (a) landscape video. Compared to (b) PWC-Net, the motion fields generated by (c) VideoFlow demonstrate superior overall quality and sharper edge preservation.}
\label{fig:videoflow_ablation}
\end{figure}

\subsection{Sketch-guided Motion Prediction}
\label{subsec:motion_field_predict}
\textbf{\textit{Motion Sketches}}. In contrast to structural sketches for the semantic structure in image generation, motion sketches control the movement of flow elements within the image. For inference, the motion sketches are created by resampling user-drawn lines to 20 sampling points, followed by a smoothing process. A gradient color from white to black is applied to the polyline vector for motion direction indication.

\textbf{\textit{Latent Motion Diffusion Model}}. To involve motion dynamics in the fluid elements of the landscape image, we introduce the latent motion diffusion model (LMDM) to predict a motion field from the generated realistic landscape image, conditioned on user-provided motion sketches and text prompts. The structure of the LMDM is shown in Fig. \ref{fig:workflow}(b). Initially, we train a CNN-based autoencoder for encoding the motion field into latent space, thereby adopting the LDM to the motion field generation task. To facilitate efficient control over motion generation via motion sketches, we train a motion ControlNet to encode sketches (Fig. \ref{fig:dataset}(c)) into guidance that determines the flow direction of each pixel within the motion field. This choice stems from ControlNet’s strong pretrained ability to interpret sketch structures. We extend this capability from sketch-to-image to sketch-to-motion: the motion ControlNet converts sparse motion strokes into dense pixel-wise guidance for determining flow directions. This enables (a) reuse of pretrained sketch understanding without training from scratch, (b) dense and fine-grained spatial control, and (c) motion that strictly follows the sketched trajectories while preserving spatial coherence. Simultaneously, to use the generated realistic landscape image and the text prompt as conditions, it is essential to project the referenced realistic landscape image into a shared embedding space that aligns semantically with the text embeddings. Inspired by IP-Adapter \cite{ye2023ipadapter}, we add a new cross-attention layer for each layer in the original U-Net model to incorporate image features (Fig. \ref{fig:dataset}(a)). The semantic control is injected as follows:
\begin{align}
\label{eqn:cross_attention}
Z&=\text{Softmax}(\frac{\mathbf{Q}\mathbf{K_{t}}^T}{\sqrt{d}})\mathbf{V_{t}} + \text{Softmax}(\frac{\mathbf{Q}\mathbf{K_{r}}^T}{\sqrt{d}})\mathbf{V_{r}}
\end{align}
where $K_t$ and $V_t$ are the key and value matrices from the text features, $K_r$ and $V_r$ are the key and value matrices from the image features, $d$ denotes feature dimension, $Z$ is the output of cross-attention layer. The image cross-attention and text cross-attention use the same query $\mathbf{Q}$.

\textbf{\textit{Dataset Preparation}}. We trained the LMDM on the same landscape dataset in \cite{holynski2021animating}, which contains 4,750 paired videos and ground truth motion fields generated by PWC-Net \cite{sun2018pwc}. For estimating higher-quality motion fields from videos, we adopted VideoFlow \cite{shi2023videoflow}, a state-of-the-art multi-frame optical flow generation framework, to regenerate the ground truth motion fields. As shown in Fig. \ref{fig:videoflow_ablation}, the results demonstrate that VideoFlow produces motion fields of higher quality, with clearer and more defined boundaries than those generated by PWC-Net. This improves the overall quality of the dataset, enabling our model to generate higher-quality motion fields.
Similar to \cite{mahapatra2023}, we used BLIP2 \cite{li2023blip} to generate the captions of the first frame of each video, as the video content undergoes minimal changes. The ground truth motion sketches consist of streamlines extracted from the motion fields. The streamlines depict the movement direction of motion fields at any moment, which provides an intuitive visualization approach to represent the geometric structure and dynamics of motion fields. 
Fig. \ref{fig:dataset} presents examples from the dataset, including the original landscape image, manually annotated motion sketches overlaid on the fluid region, and the corresponding motion field of the original image.

\begin{figure}[t]
\centering
\includegraphics[width=\linewidth]{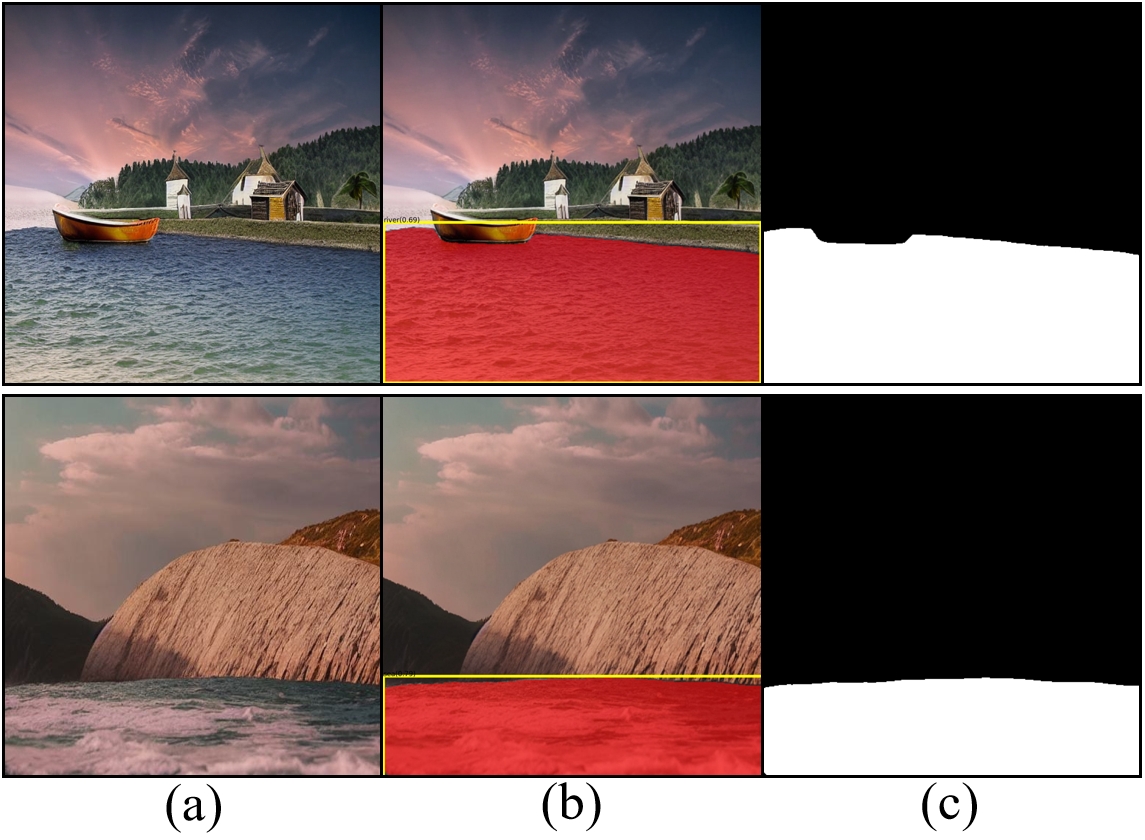}
\caption{Two-stage fluid mask extraction results (c) extracted from landscape images (a) using the bounding box (b) as intermediate detection results using Grounded SAM model. }
\label{fig:mask_extract}
\end{figure}

\begin{figure}[t]
\centering
\includegraphics[width=\linewidth]{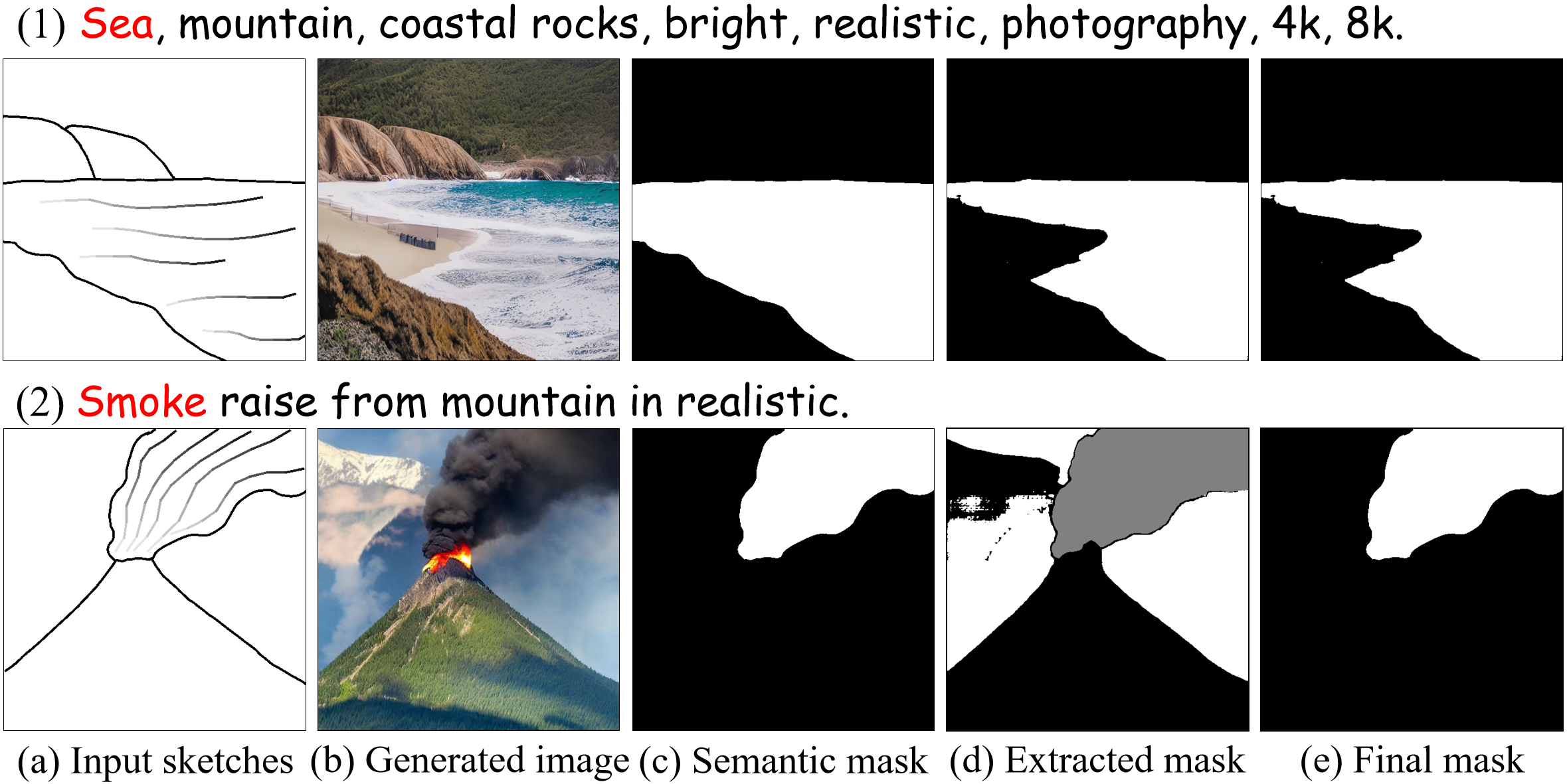}
\caption{The final fluid mask (5) is obtained by intersecting the semantic mask (3) with the extracted mask (4), preserving sketch-guided structure and boundary accuracy while removing unintended fluid regions.}
\label{fig:mask_merge}
\end{figure}

We extract the fluid mask from the generated landscape image to generate dynamic fluid elements while maintaining a static background. We observes that the input sketches contain semantic layout information. Specifically, we first perform connected-component segmentation on the structural sketch to partition the scene into a set of closed semantic candidate regions. Then, we detect all non-background stroke coordinates from the motion sketch and determine whether each structural region should be regarded as a fluid region based on the spatial coverage of motion strokes: if a region is intersected or overlapped by motion strokes, it is labeled as a fluid region, as shown in Fig. \ref{fig:mask_merge}(c). However, the semantic mask should not be regarded as a complete or perfectly accurate fluid mask. This is primarily because, in the landscape image synthesized from the sketches, the true boundaries of the fluid regions do not always strictly correspond to the contour lines drawn in the sketch. As illustrated in Fig. \ref{fig:mask_merge}(1,b), the actual sea area is noticeably smaller than the region enclosed by the sketch strokes, leaving certain coastal areas exposed. Therefore, it is necessary to extract the precise fluid regions directly from the generated landscape image. To automatically extract fluid masks with accurate edges from the generated landscape images, we utilize the Grounded Segment Anything Model \cite{ren2024grounded} with text prompt, which integrates two large language models (LLMs): (1) Grounding DINO \cite{liu2023grounding}, an approach for detecting bounding boxes of arbitrary objects based on given text queries; (2) Segment Anything Model \cite{kirillov2023segany}, which accepts the bounding boxes as input for generating a precise mask. Fig. \ref{fig:mask_extract} demonstrates the robust capability of the Grounded Segment Anything Model in accurately detecting text-guided fluid regions. We further observed an additional issue: the model may synthesize unintended fluid regions that fall outside the user-specified motion areas. As illustrated in Fig. \ref{fig:mask_merge}(2,b), the model generates smoke in the background, even though it was not intended to be animated. However, the Ground SAM identifies all fluid regions indiscriminately. To resolve this problem, our mask extraction step computes the intersection of two sources: the semantic mask (Fig. \ref{fig:mask_merge}(c)) derived from the input sketches and the predicted mask (Fig. \ref{fig:mask_merge}(d)). The resulting final mask (Fig. \ref{fig:mask_merge}(e)) simultaneously maintains the user-specified structural constraints, benefits from the boundary accuracy of the predicted mask, and excludes fluid regions not intended by the motion sketches.

\subsection{Stylized Cinemagraph Synthesis}
\label{subsec:cinemagraph_generation}
After the prediction of the motion field $F_{M}$ from the generated landscape image, the motion field can be used to compute the new two-dimensional position of each original pixel $P_{0}$ in the n-th frame to the next frame, where $n \in \{ 0,1,...,N \}$.
\begin{equation}
\label{eqn:euler_integration}
P_{n+1} = P_{n}+F_{M}(P_{n}), F_{M}(P_{n}) = F_{n \to n+1}(P_{n}),
\end{equation}
Similar to \cite{holynski2021animating}, we adopt the Euler integration method to reduce the number of model inference iterations,
\begin{equation}
\label{eqn:euler_integration}
F_{0\to n}(\hat{P}_0)=F_{0\to n-1}(\hat{P}_0)+F_{M}(\hat{P}_0+F_{0\to n-1}(\hat{P}_0)),
\end{equation}
where displacement field $F_{0\to n}$ enables direct pixels relocation, allowing pixels from the initial frame to be moved at their corresponding location $\hat{P_{0}}$ in the target n-th frame. However, directly warping in the pixel space results in large unknown holes at the edge of fluid regions over time.

The seamless looping cinemagraph can be synthesized with symmetric splatting in the deep space \cite{holynski2021animating}. Specifically, the displacement fields are applied to bi-directionally warp the deep feature map $D_{0}$ to $D_{-n}$ and $D_{n}$, utilizing a softmax function to determine the contributions of colliding source pixels in the target frame:
\begin{equation}
D_n(\hat{p}) = \frac{\sum\limits_{p \in \mathcal{P}} \alpha D_n(p) e^{Z(p)} + \sum\limits_{\hat{p} \in \hat{\mathcal{P}}} \hat{\alpha} D_{n-N}(\hat{p}) e^{Z(\hat{p})}}{\sum\limits_{p \in \mathcal{P}} \alpha e^{Z(p)} + \sum\limits_{\hat{p} \in \hat{\mathcal{P}}} \hat{\alpha} e^{Z(\hat{p})}}
\end{equation}
where $\alpha=1-\frac{n}{N}$ and $\hat{\alpha}=\frac{n}{N}$,  $\mathcal{P}$ and $\hat{\mathcal{P}}$ denote two sets of pixels which bidirectional map to the same destination pixel. Then, the feature map set $D=\{ D_{-N}, D_{-(N-1)}
, \dots, D_{0}, \dots, D_{N-1}, D_{N} \}$ are decoded and transformed into the pixel space, resulting in the frame set $I=\{ I_{-N}, I_{-(N-1)},...,I_{0},...,I_{N-1}, I_{N} \}$. The final output cinemagraph is composed by combining these individual frames. We adopt a U-Net-based frame generator~\cite{mahapatra2023} to conduct symmetric splatting in the feature space of the encoder component of the generator. Subsequently, the decoder component generates the RGB image from the encoded features. The frame generator is trained on the existing landscape dataset \cite{holynski2021animating}. In this work, we directly employed the pre-trained network for cinemagraph synthesis. Experiments demonstrate that this network can generate seamless cinemagraphs.

\section{Implementation Details}
We conducted all experiments with Intel i9-12900k 4.8GHz CPU and GeForce RTX A6000 GPU. The neural networks were implemented using Diffusers \cite{von-platen-etal-2022-diffusers}. LMDM was implemented based on Stable Diffusion(SD) v1.5 \cite{rombach2022high}. To adapt the diffusion model for motion field generation, we replaced AutoencoderKL with a lightweight motion Autoencoder. The encoder, comprising six convolutional layers with ReLU activation, compressed the motion field in $2 \times H \times W$ into a compact $4 \times H/8 \times W/8$ size, where $H$ and $W$ denote the height and width. The decoder can reconstruct it back to the original motion field, preserving the essential motion details.

To automate the separation of structural sketches from motion sketches within a single user input sketches, we implement a pixel-intensity thresholding algorithm. Let $I\in R^{H \times W \times 3}$ denote the user input sketches. We first identify the structural sketches, which are rendered in black color, by generating a binary mask $M_{struct}$. For every pixel $p$, $M_{struct}(p)=1$ if $I(p)=(0,0,0)$, and 0 otherwise. To mitigate potential aliasing artifacts along the stroke edges, we apply a morphological dilation operation to $M_{struct}$ with a $3 \times 3$ kernel. The final structural sketches $S$ is obtained directly from $M_{struct}$. Subsequently, the motion sketches $M$ are derived via replacing all pixels belonging to the structural mask with the background color (white), formulated as $M_{struct}(p)=1_{white}$ where $M_{struct}(p)=1$, and $M_{struct}(p)=I_{p}$ otherwise. This ensures that the white-to-black gradients representing motion dynamics are isolated without interference from structural topologies.

For extracting the streamlines from the ground truth motion field, we map the motion fields to grid-based velocity fields, with each grid containing $x$ and $y$ velocity components. The color and brightness in motion fields are converted into the direction and magnitude of velocity. The Runge-Kutta method was adopted to extract streamlines from generated velocity fields. We filter the streamlines by velocity magnitude to simplify the motion sketches while preserving the main structure of the motion field. A linear gradient ranging from white to black denotes the direction of the sketched motions.

The training of LMDM consists of two stages. In the first stage, the motion ControlNet was trained to extract motion sketch features, which were used as conditions injected into the denoising step of LMDM. To reduce the number of training epochs, we inverted the colors of the ground truth of motion sketches and continued training using the provided scribble condition weights \cite{zhang2023adding}, as well as the denoising U-Net in SD v1.5. After training the motion ControlNet, a novel cross-attention layer for image features was added to the denoising U-Net. In the second stage, the first frame of each video was encoded by the image encoder of CLIP \cite{radford2021learning}, and subsequently projected to the text embedding space via an image project model employing a Multi-Layer Perceptron (MLP) network. During training, the weights of motion ControlNet and U-Net from the previous stage were frozen, while the weights of the cross-attention layers for image features and the image projection model were updated. The paired data in the dataset were randomly cropped to $512 \times 512$ resolution. We employed the AdamW optimizer with a constant learning rate of $5 \times 10^{-6}$ for training all models.

In this paper, our generated cinemagraphs consist of 120 frames (4 seconds at 30 FPS). By enforcing alignment between the last and first frames, we ensure seamless looping behavior. This approach guarantees temporal stability, resulting in smooth animations without any noticeable flickering or discontinuities at the loop junction.

\begin{figure*}[t]
\centering
\includegraphics[width=\textwidth]{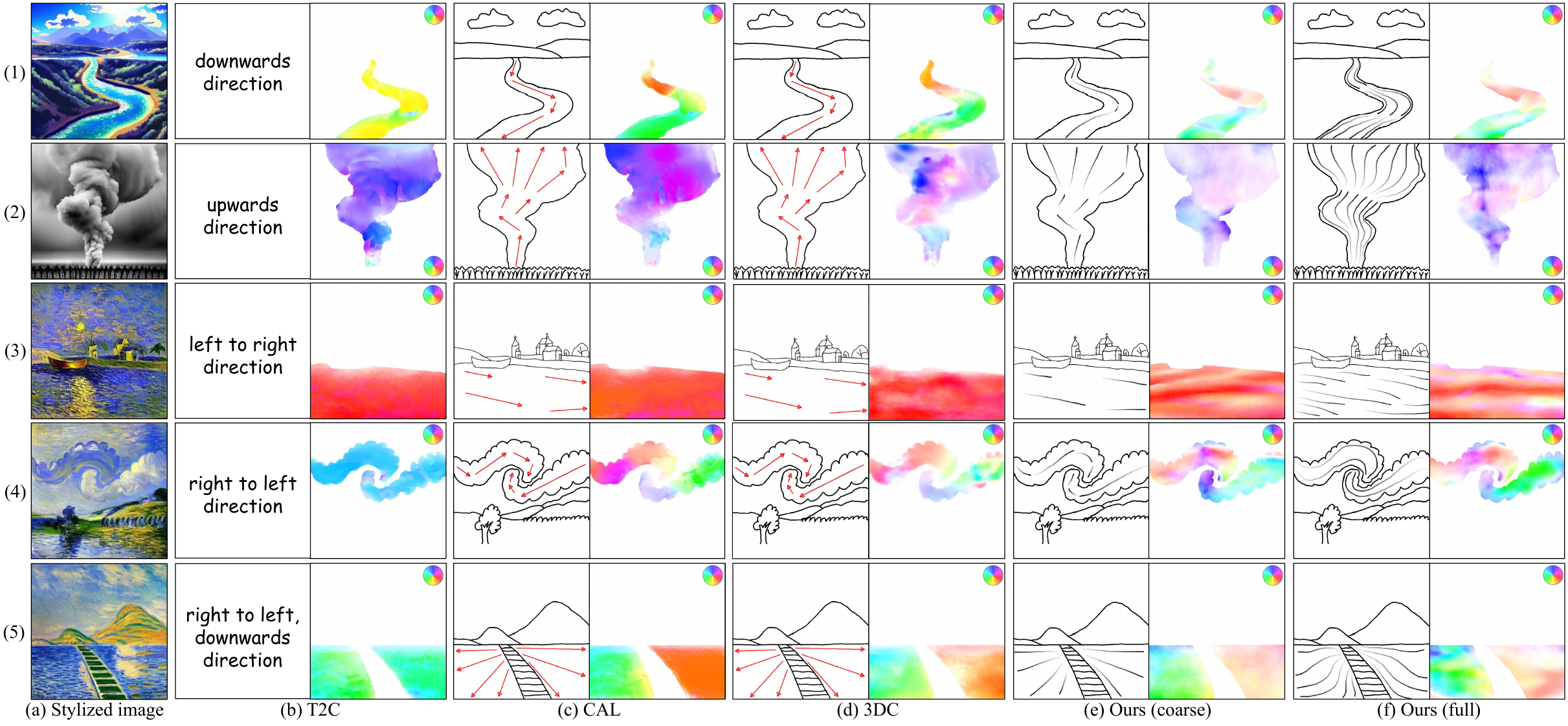}
\caption{Compared with the hint-based CAL (c), hint-based 3DC (d) and text-based T2C (b) control paradigms, the Sketch2Cinemagraph framework is capable of generating fluid motions with better temporal continuity and greater dynamic flexibility using both coarse-grained (e) and full-grained (f) motion sketches, further demonstrating the effectiveness of the proposed motion control mechanism. (\textbf{Comparisons of generated cinemagraphs are available in our supplement video.}).}
\label{fig:control_compare}
\end{figure*}

\section{Experiments and Results}
In this section, we evaluate Sketch2Cinemagraph through a series of experiments, assessing its performance from both qualitative and quantitative perspectives. The visual comparison of the generated cinemagraph results is provided in the supplementary video.

\begin{figure}[t]
	\centering
    \includegraphics[width=\linewidth]{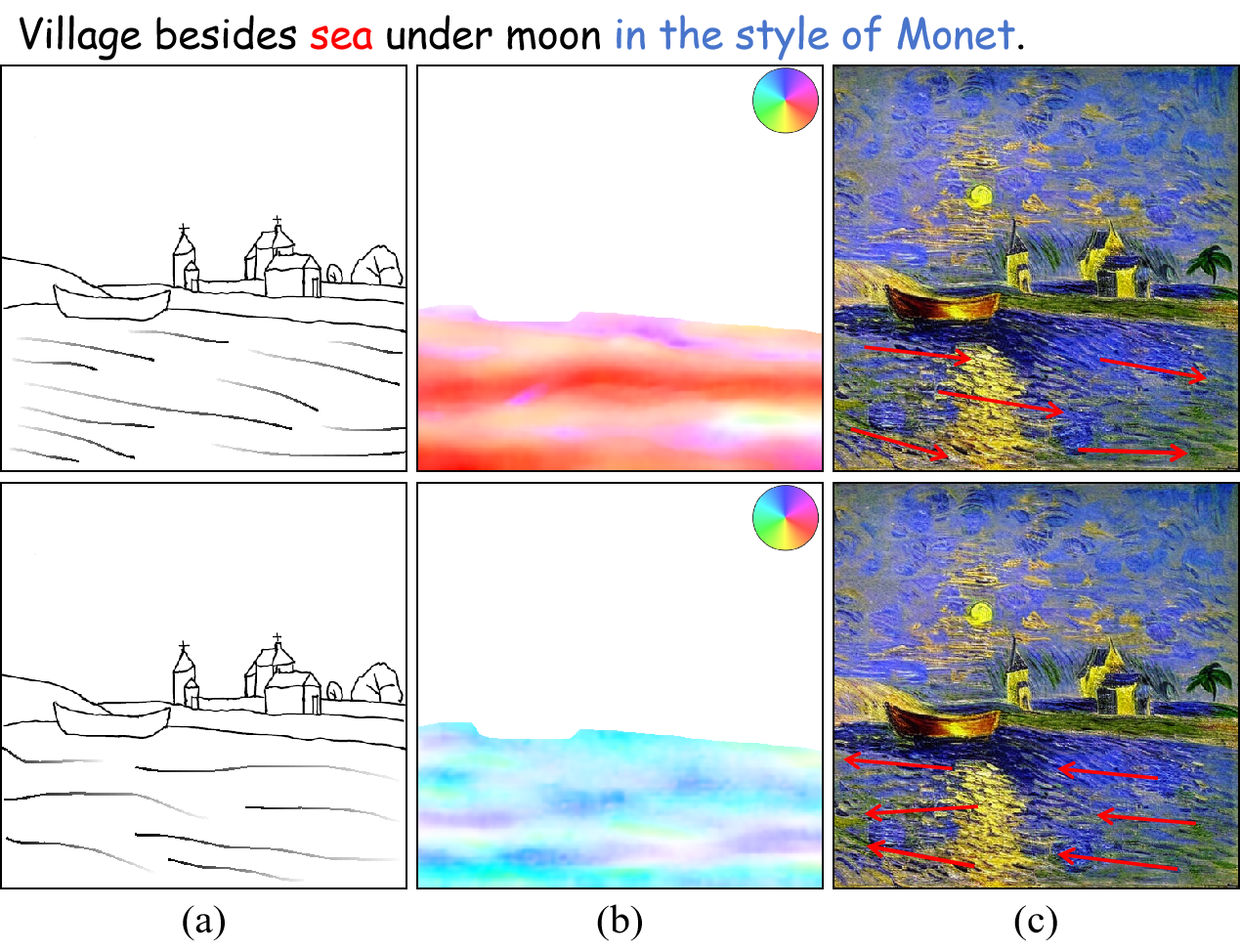}
	\caption{Example of motion field generation (b) with different motion sketches (a) for the same stylized cinemagraph (c). (The generated cinemagraphs are available in our supplementary video.)}
	\label{fig:change}
\end{figure}

\subsection{Sketch Control}
We conduct a comparative experiment of control methods using examples such as ``(1) meandering rivers'', ``(2) dynamically evolving smoke vortices'', ``(3) turbulent ocean waves'', ``(4) churning clouds'' and ``(5) multi-region sea'' in Fig. \ref{fig:control_compare}. Compared to the unidirectional control of Tex2Cinemagraph (T2C) and the motion hints of CAL \cite{mahapatra2022controllable} and 3D Cinemagraph (3DC) \cite{li20233d}, our method exhibits more flexible and seamless motion generation with both coarse and full motion sketches, demonstrating superior performance in handling fluids with intricate and variable flow characteristics, especially in regions with sharp curvature changes. T2C can only generate simple cinemagraph motion because it relies on text to specify a single direction, such as ``upwards/downwards" and ``left/right to right/left" direction. This constrains T2C’s capacity to generate fluid flows exhibiting nonlinear dynamics, as shown in Fig. \ref{fig:control_compare}(b). Moreover, this method cannot independently generate motions for multiple fluid regions within landscape images. In comparison with T2C, our Sketch2Cinemagraph enables interactive and independent control of multiple fluid regions, as shown in Fig. \ref{fig:control_compare}(4,e)(5,e). CAL utilizes hints but, due to the need for interpolation and refinement, may cause misaligned flows and discontinuities in regions with flow variations, greatly reducing the naturalness of the generated fluid motion, as shown in Fig. \ref{fig:control_compare}(c). The motion produced by CAL primarily propagates along the provided linear hints, resulting in inherently straight flow trajectories. When we attempted to approximate a curved trajectory by decomposing it into multiple linear segments as CAL inputs, the synthesized motion displayed clear discontinuities and perceptibly unnatural transitions between segments, as shown in Fig. \ref{fig:control_compare} (1,c)(4,c). This limitation extends to 3DC, which employs a similar distance-based interpolation mechanism. Consequently, 3DC generates motion fields that closely resemble those of CAL and suffers from the same artifacts when handling curved flows, as shown in Fig. \ref{fig:control_compare} (1,d)(4,d). In contrast, our Sketch2Cinemagraph can generate smooth and continuous motion fields directly from complex curves using LMDM. As shown in Fig. \ref{fig:control_compare}(e), when we provide LMDM with coarse motion sketches at the same granularity as the input hints of CAL and 3DC, the resulting motion field exhibits smoother and more continuous characteristics in regions with sharp flow variations compared to that generated by CAL and 3DC, leading to more seamless and realistic flow effects. When full motion sketches are provided to LMDM, its advantage becomes even more pronounced — the model excels in capturing fine-grained local details, maintaining flow boundary transitions, and preserving stability in high-curvature regions, thereby achieving a progressive motion control capability from coarse to fine, as shown in Fig. \ref{fig:control_compare}(d). Consequently, our method generates fluid motion fields with enhanced trajectory flexibility and superior continuity, effectively reproducing complex curvilinear flow patterns.

\begin{figure}[t]
\centering
\includegraphics[width=\linewidth]{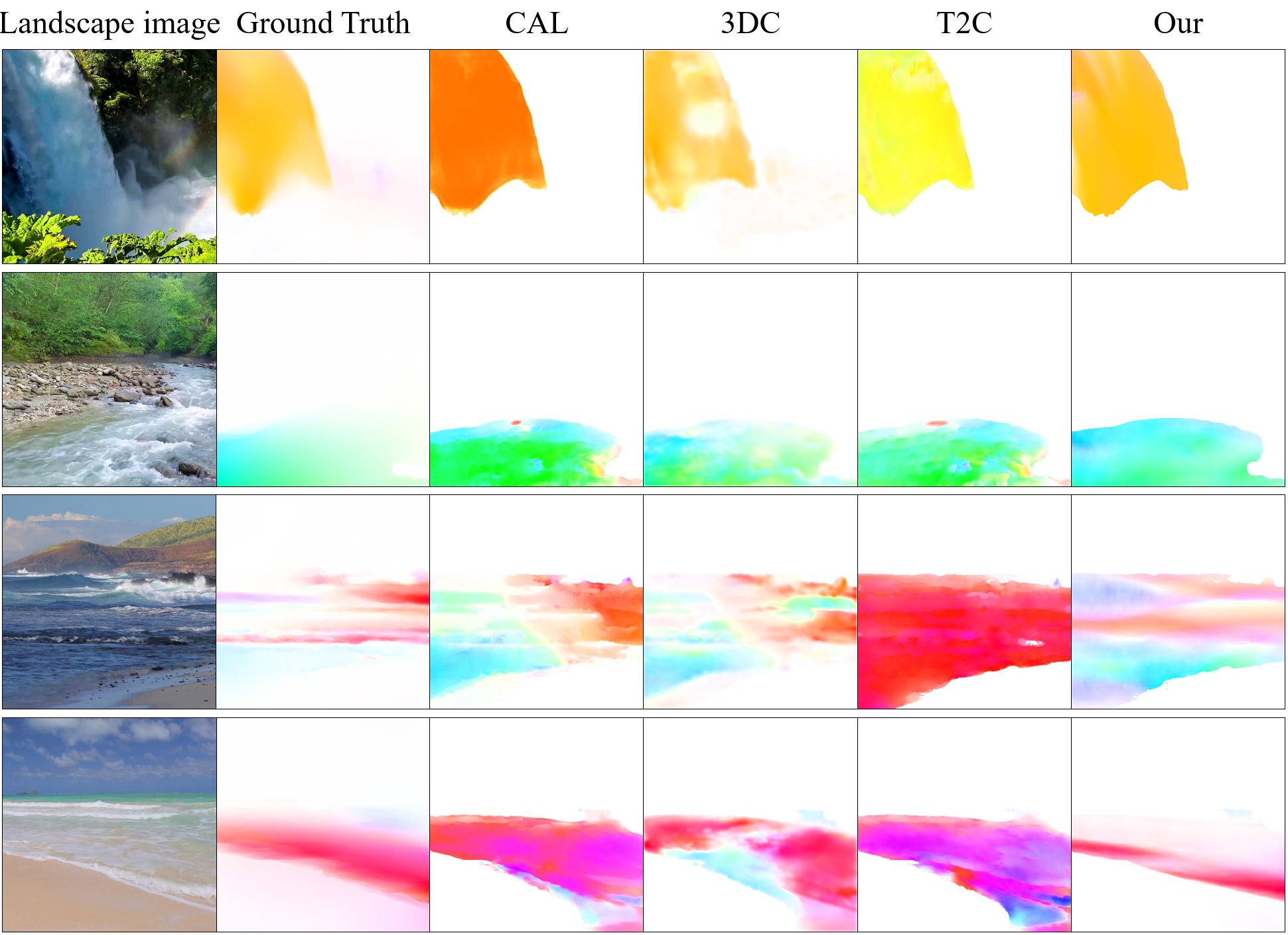}
\caption{Visual comparisons with CAL, 3DC and T2C for motion field prediction demonstrate that our Sketch2Cinemagraph generates more realistic motions that better align with the target ground truth.}
\label{fig:compare}
\end{figure}

\begin{figure}[t]
	\centering
	\includegraphics[width=\linewidth]{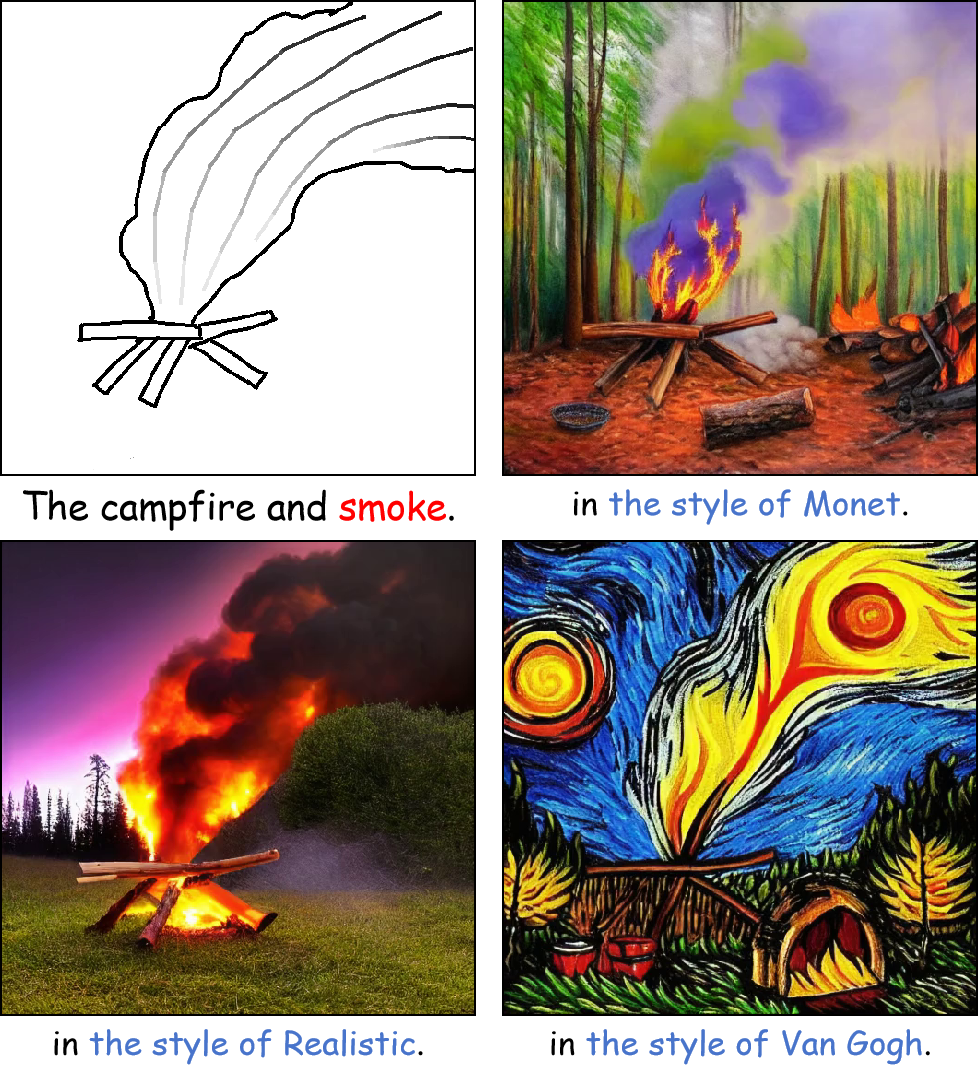}
	\caption{Various stylized cinemagraphs were generated using the same structural and motion sketches. (The generated cinemagraphs are available in our supplementary video.)}
	\label{fig:same_sketch}
\end{figure}

\begin{figure*}[t]
\centering
\includegraphics[width=\linewidth]{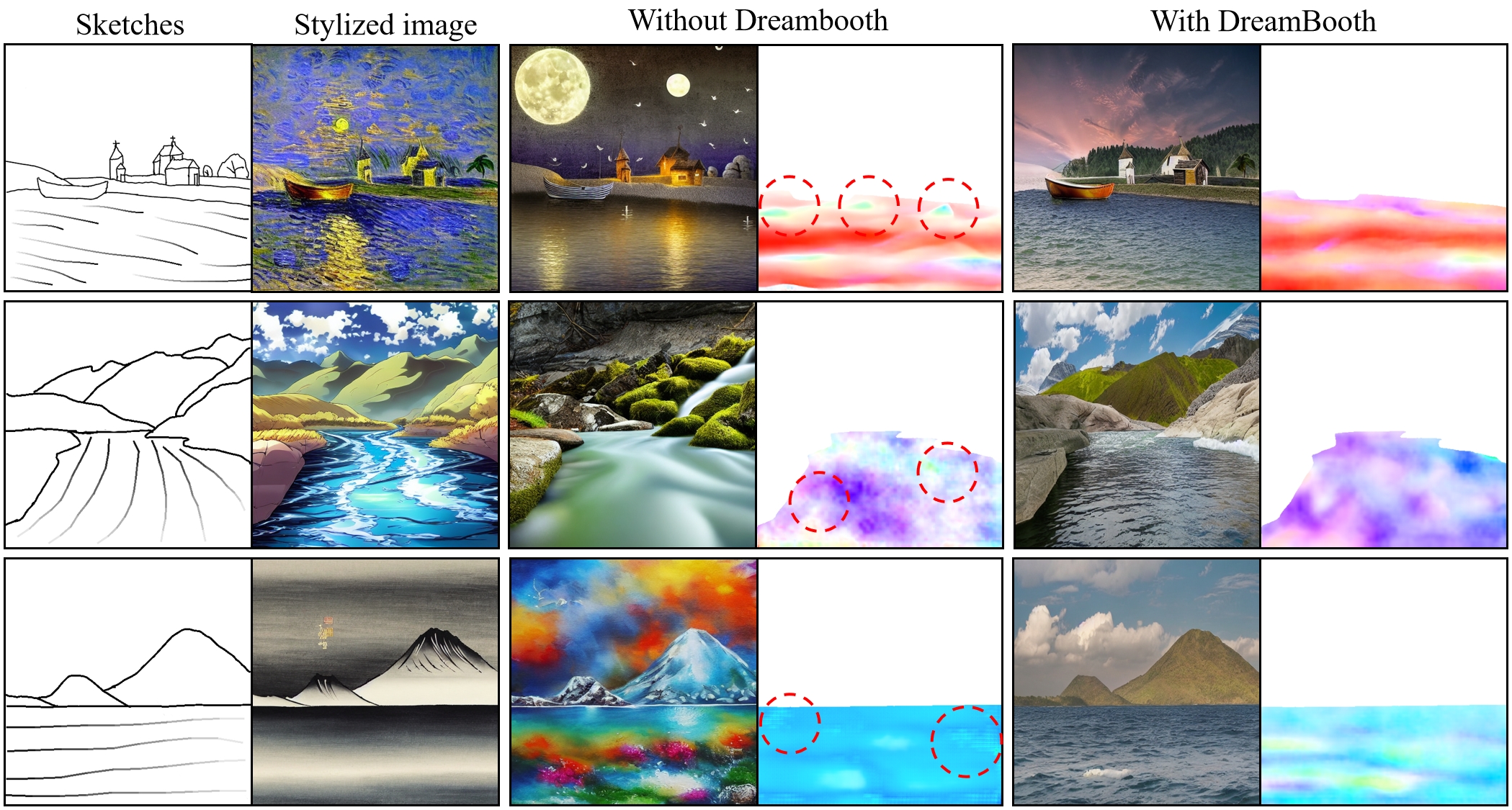}
\caption{An ablation study evaluating the impact of realistic landscape images on motion field prediction. Cinemagraphs generated with a realistic image (Fig. \ref{fig:Ablation_realistic}(a)) are compared to those generated with an unrealistic image (Fig. \ref{fig:Ablation_realistic}(b)).}
\label{fig:ablation_dreambooth}
\end{figure*}

\begin{figure}[t]
	\centering
		\includegraphics[width=\linewidth]{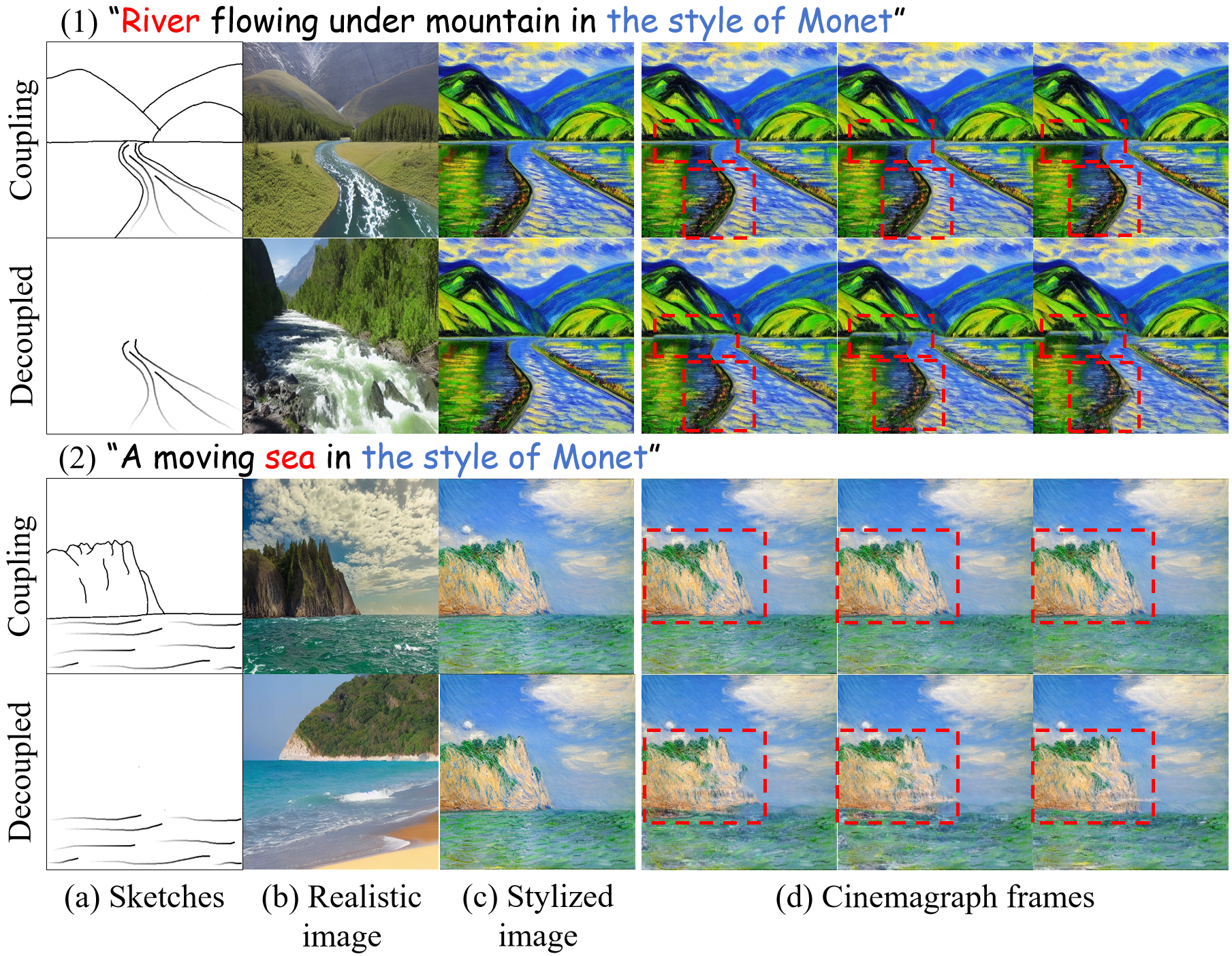}
	\caption{Comparison of motion artifacts between a decoupled baseline and our proposed coupling framework. The cinemagraph frames of the decoupled baseline exhibit significant motion bleeding, where static background elements (e.g., riverbanks, rocks) are erroneously warped along with the fluid motion due to the lack of structural alignment (highlighted with red rectangles).}
	\label{fig:coupling}
\end{figure}

\subsection{Qualitative Evaluation}
We qualitatively evaluated our sketch-guided LMDM's ability for flexible movement control of stylized cinemagraphs. Fig. \ref{fig:change} shows the motion field generated using different input motion sketches. We observed that the generated flow motion closely matched the input motion sketches, demonstrating the robustness of our LMDM model in generating motion fields that align with input motion sketches.

Similar to CAL, recent methods for controllable cinemagraph generation \cite{li20233d, choi2024} convert sparse hints into dense motion fields. Therefore, we compare our sketch-guided LMDM with CAL and 3DC regarding motion quality. Since hint-based CAL and 3DC rely on different control conditions than our method, we extract streamlines and five hints (CAL and 3DC's best practice) from the ground truth motion field to ensure an objective comparison under identical motion conditions. The extracted hints and streamlines are used as motion control conditions for the CAL, 3DC and our Sketch2Cinemagraph method during evaluation. This consistency helps reduce result bias caused by differences in control conditions. In addition, we compare our sketch2Cinmagraph with Text2Cinemagraph, which enables the generation of cinemagraphs with text-guided motion direction control. To ensure the fluid region is the same, we used the fluid mask extracted from the real motion field as the constraint region for T2C, CAL, 3DC and our Sketch2Cinmagraph. The comparison of the generated motion fields is shown in Fig. \ref{fig:compare}. The motion fields generated by our Sketch2Cinemagraph closely matched the ground truth. Fig. \ref{fig:same_sketch} shows ``campfire and smoke'' cinemagraphs in various styles generated from the same structural and motion sketches. It demonstrated that our model performs well not only in stylized domains but also in realistic domains.

\subsection{Quantitative Evaluation}
To evaluate the quality of the generated motion field, we choose Peak Signal-to-Noise Ratio (PSNR), Multi-Scale Structural Similarity (MS-SSIM) \cite{wang2003multiscale}, Average Endpoint Error (AEPE) \cite{barron1994performance}, and Mean Squared Error (MSE) to measure the similarity between the generated motion fields and the ground truth. Table \ref{tab:compare} shows that our generated motion fields more closely aligned with the target ground truth than CAL, 3DC and T2C methods. Higher PSNR and MS-SSIM scores reflect improved structural alignment, while lower AEPE and MSE values indicate more accurate motion estimation in terms of flow magnitude and direction.

\begin{table}[h]
\centering
\caption{Quantitative comparisons regarding motion field quality with ground truth regarding PSNR, MS-SSIM, AEPE, and MSE.}
\begin{tabular}{|c|c|c|c|c|}
\hline
Method & PSNR↑            & MS-SSIM↑        & AEPE↓           & MSE↓            \\ \hline
T2C    & 15.2333          & 0.7481          & 0.3227          & 0.2103          \\
CAL    & 16.8566          & 0.8053          & 0.2589          & 0.1607          \\
3DC    & 19.8456          & 0.8311          & 0.2561          & 0.1572          \\
Ours   & \textbf{21.4019} & \textbf{0.8440} & \textbf{0.2532} & \textbf{0.1557} \\ \hline
\end{tabular}

\label{tab:compare}
\end{table}

\begin{figure}[t]
\centering
\includegraphics[width=\linewidth]{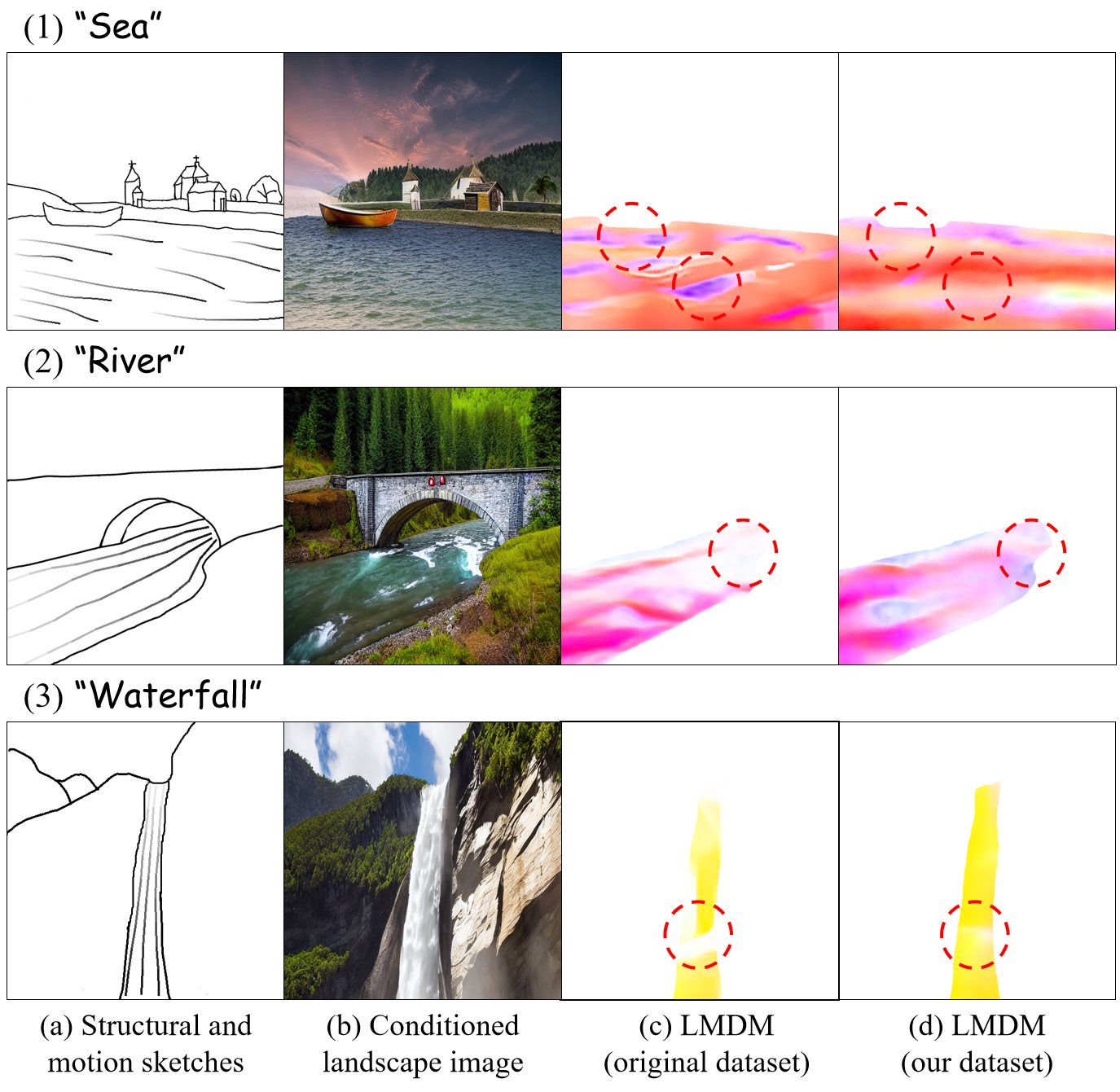}
\caption{Ablation on ground-truth motion field generation using VideoFlow. Compared with LMDM trained on the original dataset (motion fields estimated by PWC-Net), LMDM trained on the enhanced dataset (motion fields estimated by VideoFlow) produces more temporally coherent motion fields, especially around high-frequency flow regions and occlusion boundaries.}
\label{fig:pwc_videoflow_ablation}
\end{figure}

\begin{figure}[t]
\centering
\includegraphics[width=\linewidth]{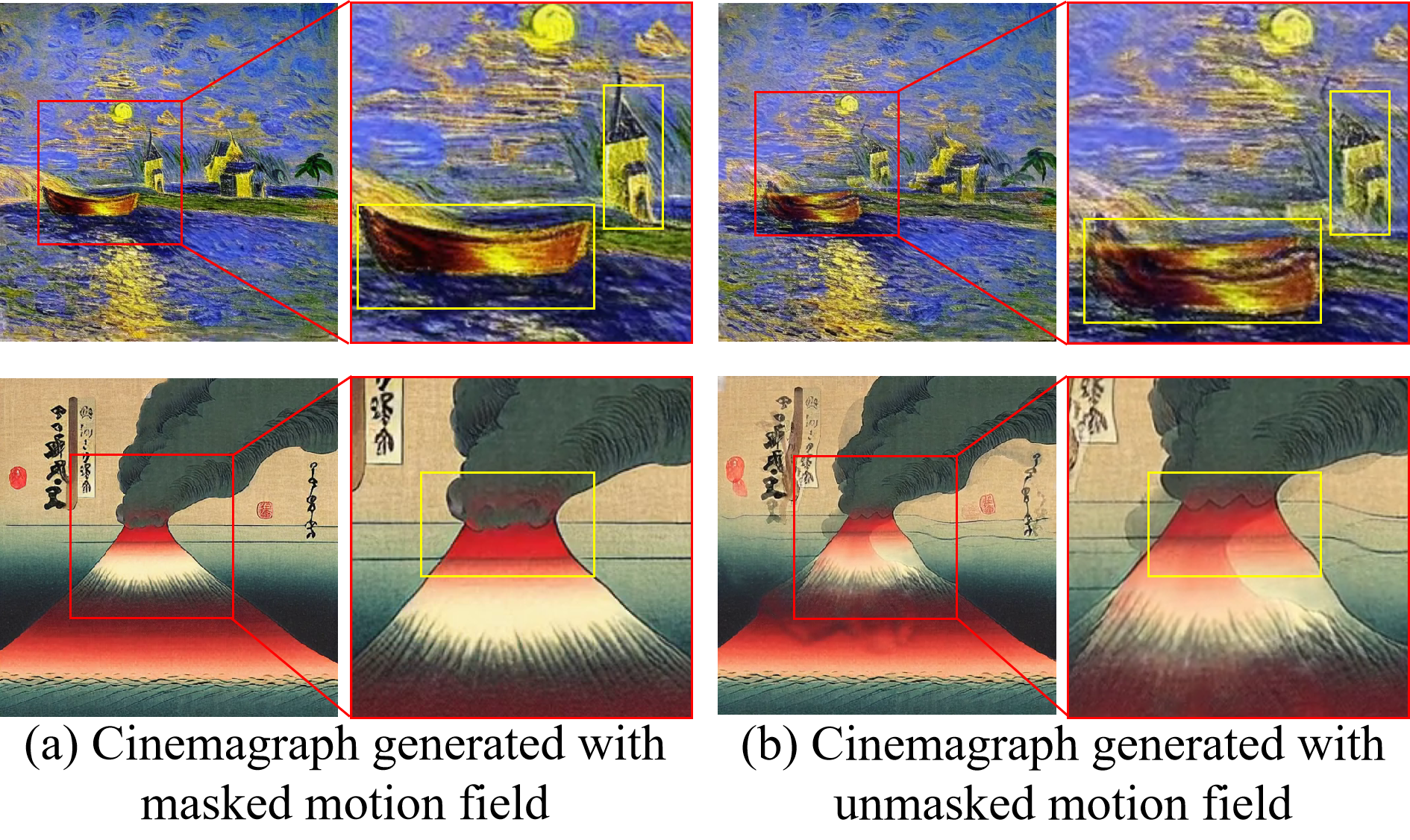}
\caption{Ablation Study (fluid mask extraction). Comparing with (b) the frame directly wrapped with unmasked motion field, (a) the frame wrapped with masked motion field can maintain the static appearance of non-fluid regions.}
\label{fig:mask_ablation}
\end{figure}

\begin{table}[h]
\centering
\caption{Quantitative comparisons cinemagraphs with ground truth videos regarding FVD, LPIPS and VMAF.}
\begin{tabular}{|c|c|c|c|}
\hline
Method & FVD↓             & LPIPS↓   & VMAF↑       \\ \hline
T2C    & 1884.82          & 0.2418   & 34.3197       \\
CAL    & 1616.76          & 0.1659   & 39.4753        \\
3DC    & 1546.37          & 0.4053   & 5.2029       \\
Ours   & \textbf{1535.99} & \textbf{0.1567} & \textbf{39.6215} \\ \hline
\end{tabular}
\label{tab:video_compare}
\end{table}

We evaluate the generated cinemagraphs against ground-truth videos using Frechet Video Distance (FVD) with the pre-trained I3D \cite{szegedy2016rethinking} model , LPIPS \cite{zhang2018unreasonable} with the AlexNet model and VMAF \cite{rassool2017vmaf}. Table \ref{tab:video_compare} shows that our method achieves lower FVD and LPIPS scores and higher VMAF scores than CAL, 3DC and T2C, indicating that it more closely preserves fluid motion characteristics and visual fidelity with respect to the ground-truth landscape videos.

\subsection{Ablation Study}
To assess the impact of our design choices in the proposed method, we conducted a series of ablation studies, specifically focusing on the effectiveness of structure-motion coupling mechanism, ground truth motion field generated by VideoFlow, fluid mask extraction and the generated realistic landscape images in sketch-guided motion field prediction stage.

\textbf{\text{Structure-motion coupling mechanism}}. We verify the structure-motion coupling mechanism by comparing it against a decoupled baseline in which the motion module is operated on realistic images without shared structural constraints. Fig.~\ref{fig:coupling} highlights the resulting misalignment: because the fluid motion field is inferred from an unconstrained realistic layout, it fails to match the semantic boundaries of the stylized image. This mismatch causes motion bleeding, leading to unnatural deformations in static regions such as riverbanks Fig.~\ref{fig:coupling}(1, d) and rocks Fig.~\ref{fig:coupling}(2, d). Our method resolves this by using the structural sketch as a geometric lock, which forces the generated appearance and motion field to align precisely, preserving the stability of the background.

\textbf{\text{Ground truth motion field generated by VideoFlow}}. To demonstrate the advantages of replacing PWC-net with VideoFlow for ground truth motion-field generation, we also trained LMDM on the original dataset in which motion fields were estimated by PWC-Net. As shown in Fig. \ref{fig:pwc_videoflow_ablation}, the LMDM trained with PWC-Net–based motion fields tends to produce discontinuous motion, local jittering, and unstable flow directions. In contrast, the LMDM trained with VideoFlow-based motion fields produces smoother and more coherent fluid motions.

\begin{figure}[t]
	\centering
		\includegraphics[width=\linewidth]{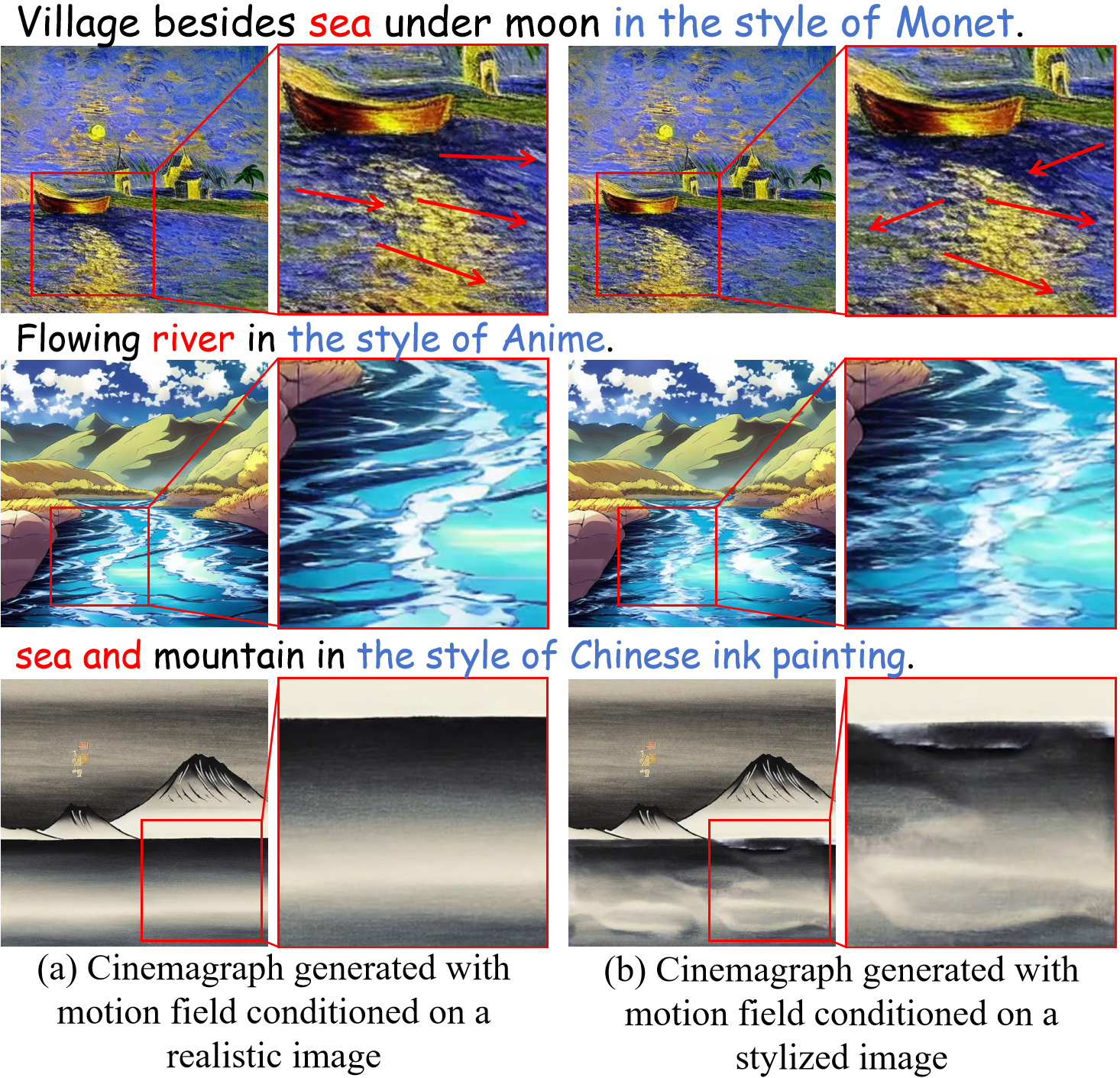}
	\caption{Cinemagraphs generated using motion fields conditioned on realistic landscape images (a) exhibit more natural water motion than those conditioned on stylized landscape images (b).}
	\label{fig:Ablation_realistic}
\end{figure}

\textbf{\text{Masked/Unmasked motion field}}. We compare the frames generated with masked and unmasked motion fields. The frame warped using unmasked motion field shows distortions in areas that should remain static, as shown in Fig. \ref{fig:mask_ablation} (b). In contrast, the masked motion field effectively restricts motion to fluid regions, preserving the integrity of static areas Fig. \ref{fig:mask_ablation} (a). This demonstrates the high accuracy of the mask boundaries extracted by Grounded SAM.

\textbf{\text{Realistic landscape image conditioning}}. The realistic images are generated using a fine-tuned LDM with DreamBooth for motion fields prediction. We conduct an ablation study on motion field generation, comparing results using realistic images generated by the fine-tuned LDM and the original LDM as shown in  Fig. \ref{fig:ablation_dreambooth}. When using images generated by the original LDM without DreamBooth fine-tuning as input, the resulting motion fields maintain the overall direction but lack detailed flow variations, exhibiting discontinuities and erroneous regions, highlighted by red circles in Fig. \ref{fig:ablation_dreambooth}. The comparison of generated cinemagraphs are shown in Fig. \ref{fig:Ablation_realistic}. This highlights the effectiveness of DreamBooth in producing realistic landscape images for motion field prediction. 

\begin{figure}[t]
	\centering
        \includegraphics[width=\linewidth]{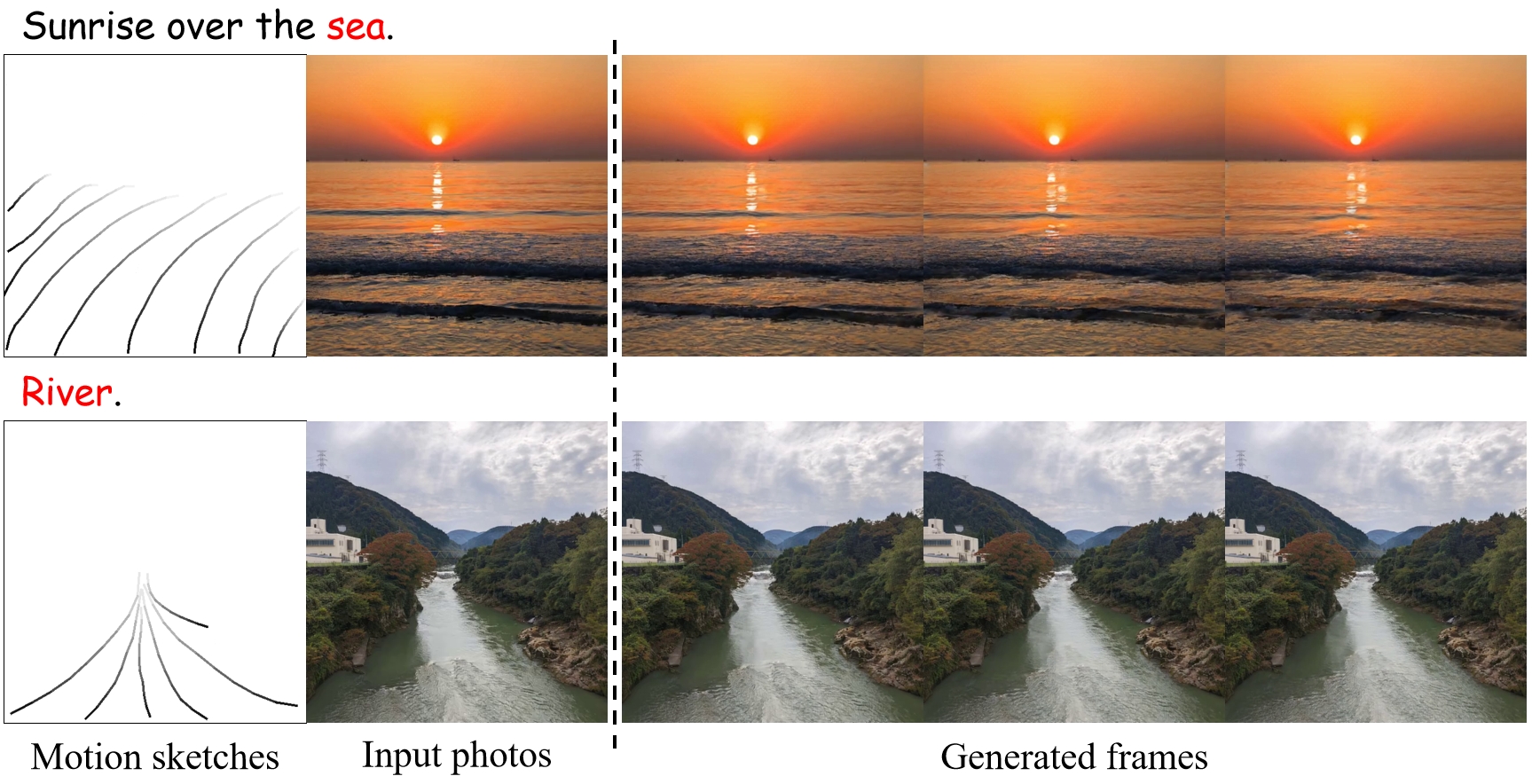}
	\caption{The examples of animated results of real-world scenes.}
	\label{fig:photo_anime}
\end{figure}

\subsection{Preference Study}
To evaluate our proposed Sketch2Cinemagraph in comparison with T2C, CAL and 3DC, we conducted a subjective preference study with 30 participants aged between 23 and 30. They were asked to evaluate visual and motion qualities on 5-point Likert scale (1 for very poor, 5 for very excellent). To ensure that participants evaluated motion quality with respect to semantic correctness rather than personal aesthetic preference, all participants were also provided with the textual prompts used to generate the cinemagraphs. This allowed them to judge whether each fluid type (e.g., river, sea waves, waterfall) exhibited dynamics that were consistent with its described behavior in the prompt. The statistical results of the preference study are summarized in Table \ref{tab:user_study}. It is verified that our method significantly outperformed previous approaches in generating cinemagraphs with high-quality motion. In addition, 83\% of the participants agreed that our motion sketches can produce better fluid motions, while 60\% preferred using sketches for cinemagraph synthesis. This highlights the superiority of our cinemagraph synthesis method in enhancing user experience.

\begin{table}[h]
\caption{A user preference study of generated stylized cinemagraphs assessing overall visual and motion qualities.}
\centering
\begin{tabular}{|c|c|c|}
\hline
Methods &  Visual Quality & Motion Quality  \\ \hline
T2C  & 3.133 $\pm$ 0.991     &  2.500 $\pm$ 1.155 \\
CAL & 2.733 $\pm$ 1.070       & 2.717  $\pm$ 0.923 \\
3DC & 1.875 $\pm$ 1.144       & 2.450  $\pm$ 1.102 \\
Ours & \textbf{3.925 \( \pm \) 0.923}  & \textbf{4.042 \( \pm \) 0.970} \\
\hline
\end{tabular}
\label{tab:user_study}
\end{table}

\subsection{Image-based Cinemagraph Synthesis}
The proposed Sketch2Cinemagraph is also applicable to image-based animation tasks, such as animating photographs captured from the real world. Fig. \ref{fig:photo_anime} showcases the animated photographs of the river and sunrise over the sea. In the river scene, the detailed structures and flow directions of the ripples are precisely preserved, while in the sunrise scene, the subtle undulations of the sea surface and the dynamic effects of light reflections are faithfully captured. The natural and lifelike fluid motions highlight the effectiveness of our method in animating real-world landscape scenes. This generalization capability stems from the inherent modularity of our framework. By decoupling sketch-guided motion field prediction from the landscape image generation process, Our LMDM can directly generate motion fields from input motion sketches and real-world landscape images. Consequently, it robustly handles real-world photography, extending the applicability of our method besides the synthesized domain. This feature significantly expands the utility of our framework, allowing users to animate static assets like travel photos or historical images. Through precise sketch-guided control, users can introduce fluid motion while maintaining the strict realism of the original input.

\section{Conclusions}
This paper proposed Sketch2Cinemagraph, a sketch-guided generation framework for stylized landscape cinemagraphs from freehand sketches. This method can generate visually appealing cinemagraphs based on user-provided structural and motion sketches with intuitive control. This approach enables amateur users without design skills to create landscape cinemagraphs and makes cinemagraph creation accessible to a broader audience. Through evaluation experiments, we demonstrated that our method outperformed all baseline approaches. For limitations, our approach may be unsuitable for simulating realistic fluids similar to other cinemagraph synthesis works \cite{holynski2021animating,mahapatra2023}. 
During the landscape image generation, the pre-trained latent diffusion model may generate extraneous objects in the fluid regions, such as rocks or ships. To address this issue, alternative seed values are required to generate images that accurately match the flow structure. The code and trained models will be publicly released in the future to support reproducibility and further research.

{
    \small
    \bibliographystyle{ieeenat_fullname}
    \bibliography{main}
}

\clearpage
\setcounter{page}{1}
\maketitlesupplementary

\section{Motion Field Visualization}
\begin{figure}[h]
\centering
\includegraphics[width=0.8\linewidth]{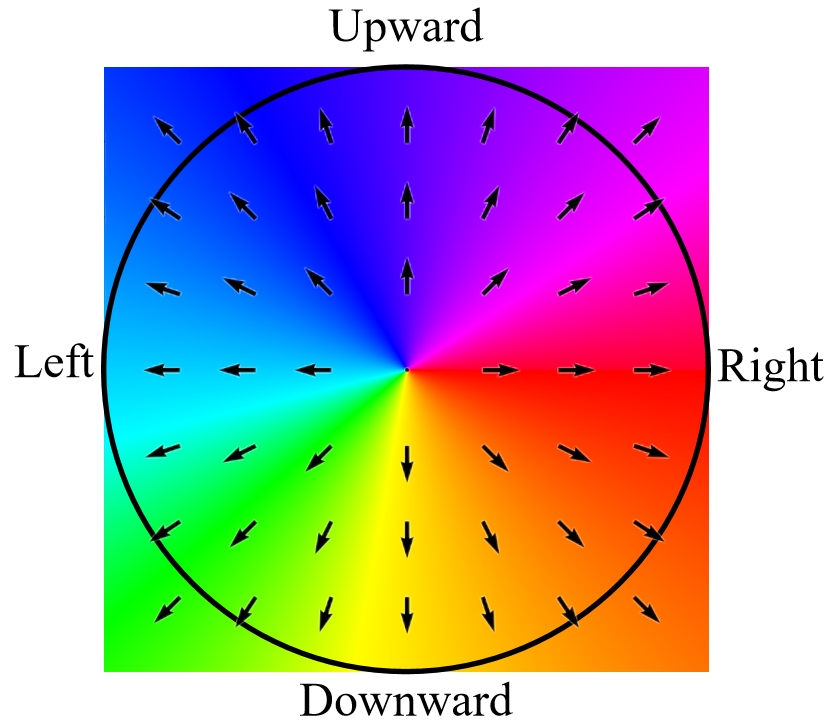}
\caption{The color wheel of motion field visualization.}
\label{fig:visual_motion}
\end{figure}

The motion field visualization method used in this paper follows the approach presented by Baker et al. \cite{baker2011database}. Fig. \ref{fig:visual_motion} illustrates the color-coding of different motion directions for each pixel in the motion fields, providing an intuitive visualization of flow patterns. Each color corresponds to a specific direction, clearly showcasing pixel-level dynamics and making the generated motion fields more intuitive and easier to understand. The arrows in the color wheel represent the directions in the 2D plane. The intensity of the color represents the speed in that direction.

\begin{figure}[h]
\centering
\includegraphics[width=\linewidth]{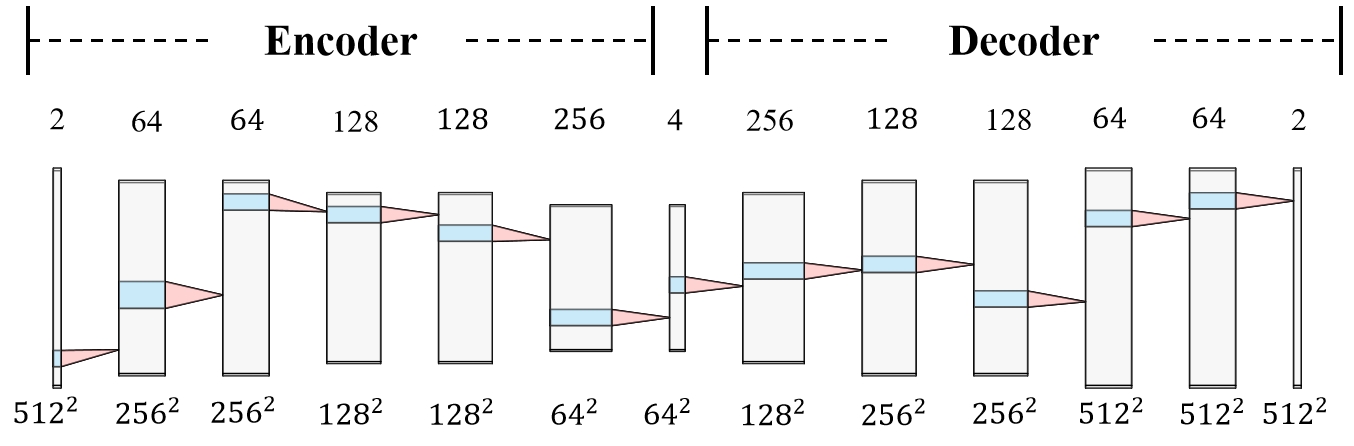}
\caption{The structure of motion Autoencoder.}
\label{fig:autoencoder}
\end{figure}

\begin{figure}[t]
\centering
\includegraphics[width=\linewidth]{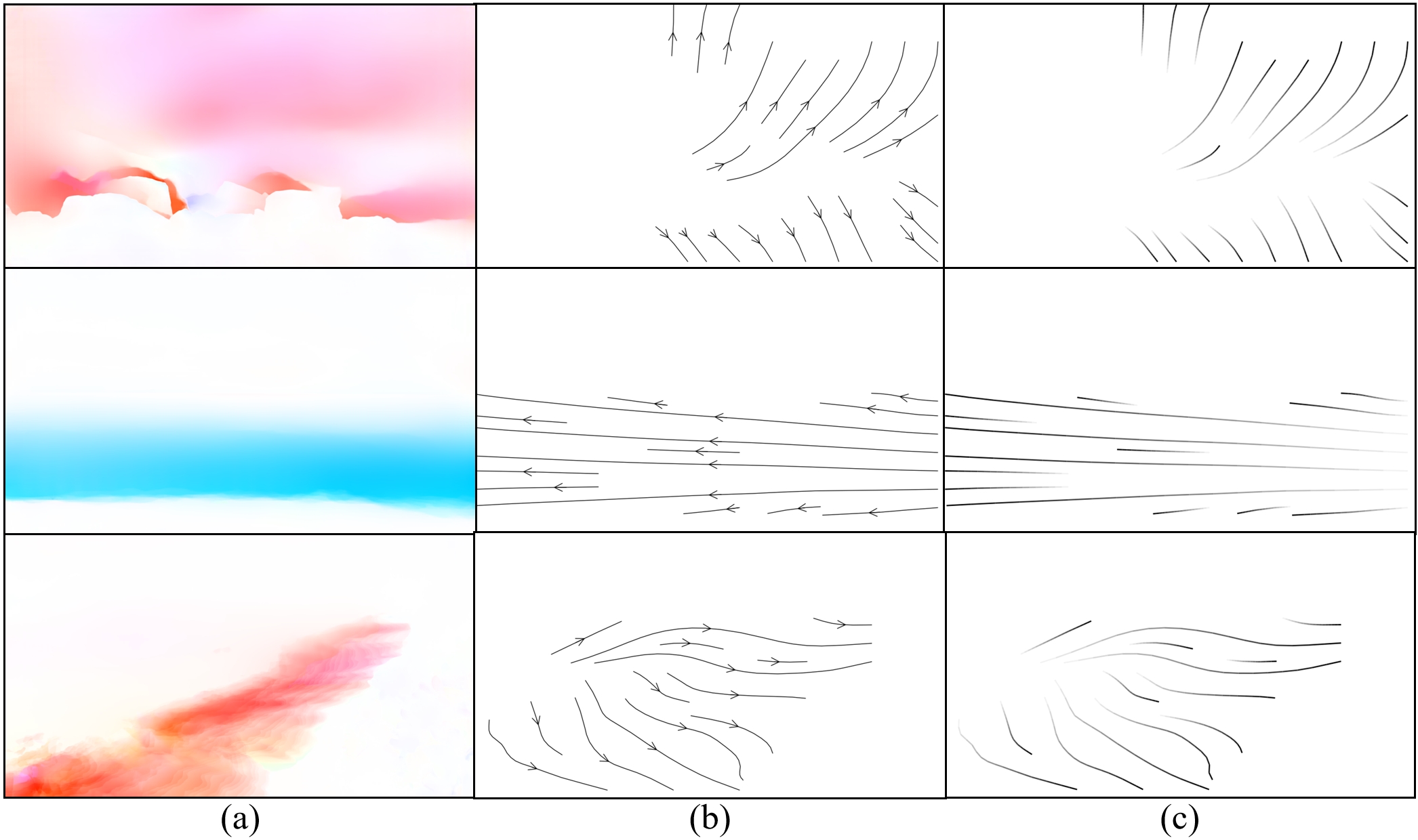}
\caption{The streamlines (a) are extracted from the motion fields and then converted into gradient gray lines to serve as ground truth motion sketches during LMDM's training (b).}
\label{fig:motion_sketch_extract}
\end{figure}

\section{Motion Autoencoder}
Fig. \ref{fig:autoencoder} illustrates the structure of the motion autoencoder in our Latent Motion Diffusion Model (LMDM). Each module of the autoencoder integrates convolutional layers and ReLU activation functions to effectively capture both spatial and temporal motion features. The motion autoencoder was trained independently, with hyperparameters including a learning rate of $lr=1 \times 10^{-4}$ and a batch size of $16$, balancing training stability and efficiency. This design and training process enables the autoencoder to generate compact and expressive latent motion representations, suitable for downstream diffusion-based synthesis tasks.

\begin{figure*}[h]
    \captionsetup{type=figure}
    \renewcommand{\thefigure}{4}
	\centering
	\begin{subfigure}{0.310\linewidth}
		\animategraphics[autoplay,loop, width=\textwidth] {20}{sec/result_folder/7/frames_with_caption_compress_1/Image}{1}{60}
	\end{subfigure}
        \begin{subfigure}{0.310\linewidth}
		\animategraphics[autoplay,loop, width=\textwidth] {20}{sec/result_folder/8/frames_with_caption_compress_1/Image}{1}{60}
	\end{subfigure}
        \begin{subfigure}{0.310\linewidth}
		\animategraphics[autoplay,loop, width=\textwidth] {20}{sec/result_folder/9/frames_with_caption_compress_1/Image}{1}{60}
	\end{subfigure}


	\begin{subfigure}{0.310\linewidth}
		\animategraphics[autoplay,loop, width=\textwidth] {20}{sec/result_folder/10/frames_with_caption_compress_1/Image}{1}{60}
	\end{subfigure}
        \begin{subfigure}{0.310\linewidth}
		\animategraphics[autoplay,loop, width=\textwidth] {20}{sec/result_folder/11/frames_with_caption_compress_1/Image}{1}{60}
	\end{subfigure}
        \begin{subfigure}{0.310\linewidth}
		\animategraphics[autoplay,loop, width=\textwidth] {20}{sec/result_folder/12/frames_with_caption_compress_1/Image}{1}{60}
	\end{subfigure}

	\captionof{figure}{Stylized cinemagraphs generated by our framework. (\textbf{The generated stylized cinemagraphs are embedded and better viewed using Adobe Reader.})}
	\label{fig:various}
\end{figure*}

\section{Streamlines Extraction}
The motion sketch used herein consists of streamlines extracted from given 2D motion field data. Streamlines represent the trajectory of fluid particles at a given instant in time. This offers an effective approach for characterizing a complex motion field. The Runge-Kutta method\cite{press2007numerical} is a common method to calculate the streamlines. In our study, we focus on extracting streamlines from a single velocity field, hence the equations are defined as follows:
\begin{equation}
\begin{split}
    x_{n+1} = x_n + \frac{h}{6}(k_{1x} + 2k_{2x} + 2k_{3x} + k_{4x}) \\
    y_{n+1} = y_n + \frac{h}{6}(k_{1y} + 2k_{2y} + 2k_{3y} + k_{4y}) \\
\end{split}
\end{equation}
\begin{equation}
\begin{split}
\begin{cases}
    k_{1x} = f_x(x_n, y_n)\\
    k_{1y} = f_y(x_n, y_n)\\
    k_{2x} = f_x(x_n + \frac{h}{2}k_{1x}, y_n + \frac{h}{2}k_{1y})\\
    k_{2y} = f_y(x_n + \frac{h}{2}k_{1x}, y_n + \frac{h}{2}k_{1y})\\
    k_{3x} = f_x(x_n + \frac{h}{2}k_{2x}, y_n + \frac{h}{2}k_{2y})\\
    k_{3y} = f_y(x_n + \frac{h}{2}k_{2x}, y_n + \frac{h}{2}k_{2y})\\
    k_{4x} = f_x(x_n + hk_{3x}, y_n + hk_{3y})\\
    k_{4y} = f_y(x_n + hk_{3x}, y_n + hk_{3y})\\
\end{cases}
\end{split}
\end{equation}

where $x_n$, $y_n$ are given as the particle position at status $n$, $x_{n+1}$, $y_{n+1}$ are the particle position at status $n+1$. $h$ is given the time step, $k_{1x}, k_{1y}$ are the slopes in x and y direction at start point, $k_{2x}, k_{2y}, k_{3x}, k_{3y}$ are the slopes at the middle points, $k_{4x}, k_{4y}$ are the slopes at the end point. $f_x$ and $f_y$ represent the x and y components of the velocity field. The original streamlines are visually represented using arrows, which lack prominent directional features. Therefore, this paper converts them into gradient gray lines to enhance their vector information, as shown in Fig. \ref{fig:motion_sketch_extract}.

\section{Examples of Stylized Cinemagraphs}
Fig. \ref{fig:various} showcases a variety of stylized cinemagraphs generated by our framework, demonstrating its ability to generate the dynamic motion of flowing skies and clouds under the provided motion sketches. These results emphasize the adaptability of our approach in handling various styles and motion scenarios, making it suitable for both artistic and realistic applications. 

\end{document}